%% file: main.tex
\documentclass{elsarticle}
\usepackage[utf8]{inputenc}
\usepackage[english]{babel}
\usepackage{xcolor, soul}

\renewcommand\hl[1]{#1}

\makeatletter
\newcount\SOUL@minus
\makeatother

\usepackage{framed}
\colorlet{shadecolor}{white}

\usepackage{graphicx}
\usepackage{amsfonts}
\usepackage{amssymb}
\usepackage{amsmath}
\usepackage{bbm}
\usepackage{mathtools}
\usepackage{float}
\usepackage{enumerate}
\usepackage[normalem]{ulem}

\usepackage{amsthm}
\theoremstyle{definition}
\newtheorem{definition}{Definition}[section]

\numberwithin{equation}{section}

\usepackage{chngcntr}
\counterwithin{figure}{section}

\DeclarePairedDelimiter\floor{\lfloor}{\rfloor}

\usepackage{pifont}
\usepackage[inline]{enumitem}

\usepackage{array}
\newcolumntype{P}[1]{>{\centering\arraybackslash}p{#1}}
\newcommand*\rot{\rotatebox{90}}

\renewcommand{\checkmark}{\ding{51}}
\newcommand{\crossmark}{} 

\usepackage{hyperref}
\hypersetup{
	colorlinks,
	citecolor=blue,
	filecolor=blue,
	linkcolor=blue,
	urlcolor=blue
}

\usepackage{caption}
\usepackage{subcaption}

\newcommand{\indep}{\perp \!\!\! \perp}

\DeclareMathOperator*{\argmax}{\arg\!\max}
\DeclareMathOperator*{\argmin}{\arg\!\min}

\usepackage{tikz}
\usetikzlibrary{arrows, arrows.meta}
\usetikzlibrary{decorations.text}
\usetikzlibrary{backgrounds}
\usetikzlibrary{shapes, shapes.geometric, positioning}
\tikzset{
    >=latex,
    node distance=2.5cm,
    every node/.style={draw, circle},
    every path/.style={draw, thick, -}
}

\begin{document}

\input{sections/00_frontmatter}
\input{sections/01_introduction}
\input{sections/02_definitions}
\input{sections/03_causal_discovery}
\input{sections/04_with_cycles}
\input{sections/05_with_interventions}
\input{sections/07_evalutation_tuning}
\input{sections/08_applications}
\input{sections/09_conclusions_future_work}

\bibliography{biblio}
\bibliographystyle{elsarticle-num-names}

\end{document}

%% file: sections/00_frontmatter.tex
\begin{frontmatter}

\journal{International Journal of Approximate Reasoning}

\title{A Survey on Causal Discovery: Theory and Practice}

\author[1]{Alessio Zanga}\corref{cor1}\ead{alessio.zanga@unimib.it}
\author[2]{Elif Ozkirimli}\ead{elif.ozkirimli@roche.com}
\author[1,3]{Fabio Stella}\ead{fabio.stella@unimib.it}

\cortext[cor1]{Corresponding author}

\address[1]{Department of Informatics, Systems and Communication,\\ University of Milano-Bicocca, \\Viale Sarca, 336, 20126 Milano, Italy}
\address[2]{Data and Analytics Chapter, \\ F. Hoffmann - La Roche Ltd, Basel, Switzerland.}
\address[3]{Bicocca Bioinformatics, Biostatistics and Bioimaging Centre (B4), Milano, Italy.}

\begin{abstract}
Understanding the laws that govern a phenomenon is the core of scientific progress. This is especially true when the goal is to model the interplay between different aspects in a causal fashion. Indeed, causal inference itself is specifically designed to quantify the underlying relationships that connect a cause to its effect. Causal discovery is a branch of the broader field of causality in which causal graphs are recovered from data (whenever possible), enabling the identification and estimation of causal effects. In this paper, we explore recent advancements in causal discovery in a unified manner, provide a consistent overview of existing algorithms developed under different settings, report useful tools and data, present real-world applications to understand why and how these methods can be fruitfully exploited.
\end{abstract}


\end{frontmatter}

%% file: sections/01_introduction.tex
\section{Introduction}\label{sec:introduction}

\subsection{A General Overview}
One of the mantras that is repeated in every statistical course is that {\em correlation does not imply causation}. This is also observed in several disciplines, such as economics \cite{imbens2004nonparametric}, biology \cite{cross2013identification}, computer science \cite{hill2011bayesian, pearl2018theoretical} and philosophy \cite{glymour2019review}. Following \cite{hernan2020whatif}, the main goal of a research study is often to assess the effect, if any, of an action on some outcome and not measuring a mere correlation. For example, this is true when it comes to decision making, since deciding which intervention must be taken is not straightforward and must be addressed properly to avoid any potential side effects. In order to identify and quantify a causal effect, the set of tools provided by causal discovery should be used accordingly. Here, the final goal is to decompose the total effect of an action into the causal and non-causal effects.

\emph{Causal inference}, i.e. the task of quantifying the impact of a cause on its effect, relies heavily on a formal description on the interactions between the observed variables, i.e. a casual graph. Such graphical representation is na\"{\i}ve in its concept, yet so effective when it comes to {\em explainability}. Following \cite{spirtes2000causation}, it boils down to connect a cause to an effect (outcome) by drawing arrows from the former to the latter, to obtain a qualitative description of the system under study. This is in stark contrast with black-box techniques, where predictions about an outcome are made with a pure data-driven approach. Indeed, these methods fall short both in terms of explainability and decision making, as stated in \cite{bareinboim20211OP, glymour2016causal, hernan2020whatif}. Therefore, when causality is empowered through the instrument of graphical models, it is possible to overcome the current limitations of machine learning and deep learning tools, enabling the researcher to reach a higher level of understanding.

When the causal graph is unknown, one may recover the cause-effect pairs by combining available data together with prior knowledge, whenever possible. The process of learning graphical structures with a causal interpretation is known as {\em causal discovery}. Recently, causal discovery has gained significant traction, especially when experimental data are available. However, this growth fragmented the landscape into multiple fields that differ for assumptions, problems and solutions, while aiming to the same goal. For this reason, this work summarizes the current status of causal discovery from both a theoretical and practical point of view, unifying shared concepts and addressing differences in the algorithms made available by the specialized scientific literature.

This survey is structured as follows. In Section \ref{sec:introduction}, the reader is provided with a general introduction to the causal discovery problem, along with an overview of previous works on the same topic. Section \ref{sec:definitions} is devoted to provide concepts, definitions and problems that are common across different approaches presented in the following pages. Section \ref{sec:observational} explores the first set of algorithms in the observational setting, while Section \ref{sec:cyclical} relaxes the acyclicity assumption. In Section \ref{sec:interventional}, the scope is extended to cover the experimental scenario, where multiple interactions with the system of interest are taken into account. Sections \ref{sec:evaluation} and \ref{sec:applications} report respectively on evaluation techniques and on practical applications of the discussed methodologies. Finally, Section \ref{sec:conclusions} draws conclusions about the current landscape of causal discovery.

\subsection{Related Works}
To the best of our knowledge, six different surveys on causal discovery were published from 2019 to 2022. In particular, \cite{nogueira2022methods} acted as a \emph{meta-survey} by checking the contents covered by the others concerning five topics, namely: theory, data, software, metrics and examples. A modified version of this checklist can be found in Table \ref{tab:meta_survey}, which was adapted for a direct comparison with the structure of our survey.

\begin{table}
    \centering
    \begin{tabular}{|p{65mm}|c|c|c|c|c|}\hline
        \small{Related Works} &
        \rot{\shortstack[c]{Theoretical\\Definitions}} &
        \rot{\shortstack[c]{Evaluation\\Datasets}} &
        \rot{\shortstack[c]{Evaluation\\Metrics}} &
        \rot{\shortstack[c]{Software\\Packages}} &
        \rot{\shortstack[c]{Practical\\Applications}} \\ \hline
        
        {\small A Survey of Learning Causality with Data: Problems and Methods \cite{guo2021survey} }&
        \checkmark & \crossmark & \checkmark & \crossmark & \checkmark \\ \hline
        
        \small{Causal Inference for Time Series Analysis: Problems, Methods and Evaluation \cite{moraffah2021causal}} &
        \checkmark & \checkmark & \checkmark & \crossmark & \crossmark \\ \hline
        
        \small{Review of Causal Discovery Methods based on Graphical Models \cite{glymour2019review}} &
        \checkmark & \crossmark & \crossmark & \crossmark & \checkmark \\ \hline
        
        \small{Causal Discovery Algorithms: A Practical Guide \mbox{\cite{malinsky2018causal}}} &
        \checkmark & \crossmark & \checkmark & \checkmark & \crossmark \\ \hline
        
        \small{D'ya like DAGs? A survey on structure learning and causal discovery \mbox{\cite{vowels2021d}}} & \checkmark & \checkmark & \checkmark & \checkmark & \crossmark \\ \hline
        
        \small{Causal Discovery in Machine Learning: Theories and Applications \cite{nogueira2021causal}} &
        \checkmark & \crossmark & \checkmark & \checkmark & \crossmark \\ \hline
        
        \small{Toward Causal Representation Learning \cite{scholkopf2021toward}} &
        \checkmark & \crossmark & \crossmark & \crossmark & \crossmark \\ \hline
    \end{tabular}
    \caption{Comparison of recent surveys on causal discovery in terms of covered contents.}
    \label{tab:meta_survey}
\end{table}

While every contribution provided adequate background knowledge and theoretical definitions involving the fundamental aspects of causal discovery, only a few examples \cite{glymour2019review, moraffah2021causal, nogueira2021causal} reported evaluation datasets or metrics, and just two of them listed both \cite{moraffah2021causal, nogueira2022methods}. The landscape is even more fragmented when observed from a practical point of view: only two contributions \cite{nogueira2021causal, nogueira2022methods} presented and discussed the availability of software tools to perform the described procedures, thus hindering the applicability of causal discovery to researchers approaching this topic for the first time.

In particular, the contributions from \mbox{\cite{guo2021survey, vowels2021d}} provide insights on the discovery procedure using machine learning, deep learning and reinforcement learning approaches. Authors in \cite{moraffah2021causal} tackled the problem of recovering the causal graph from time-series datasets, while \cite{glymour2019review} restricted its attention to the most famous techniques. Moreover, \cite{nogueira2021causal} presented a general survey on the topic without a proper {\em interventional} section, as for \cite{scholkopf2021toward} in the latent case. Researchers in \mbox{\cite{malinsky2018causal}} focused on a high-level overview of the methods to provide a general guide for practical applications. Finally, if the reader is interested in gaining a high-level perspective that investigates the interplay between causal inference and causal discovery, the content of  \mbox{\cite{nogueira2022methods}} is preferred, opposed to our in-depth approach exclusively focused on causal discovery.

This survey is designed in the light of the above considerations and aims to guide the inexpert reader through the forest of causal graphs to avoid common pitfalls when comparing and assessing the quality of results obtained by different causal discovery algorithms.
It is worthwhile to mention that this survey is different from those published from 2019 to 2022 with respect to both theory and practice. Indeed, existing surveys introduce theoretical aspects of causal discovery while only a few go into additional details.
Another lack of existing surveys in term of theory is that very few of them discuss the difference between observational and interventional data.
This survey has also many differences in terms of practice: {\em i})
we provide a descritpion on evaluation datasets and metrics, {\em ii}) we discuss how to tune strategies for choosing values of algorithm's hyperparameters, {\em iii}) we report on software packages, and {\em iv}) we discuss practical applications of causal discovery methods.

%% file: sections/02_definitions.tex
\section{Definitions and Notation }\label{sec:definitions}
This section gives the main definitions, concepts and assumptions on causality, together with the associated notation.  In particular, we give the definition of causal model together with the definition of causal discovery problem.

\subsection{Notation}
We denote mathematical objects with capital letters, such as random variable $X$, and collections of objects with capital boldface letters, such as set $\textbf{X}$.

\begin{definition}[Graph]
A graph $G = (\textbf{V}, \textbf{E})$ is a mathematical object represented by a tuple of two sets: a finite set of vertices \textbf{V} and a finite set of edges $\textbf{E} \subseteq \textbf{V} \times \textbf{V}$. If not specified otherwise, this graph is intended as an \emph{undirected graph}, where the \emph{undirected edge} $(X, Y)$ is identical to the edge $(Y, X)$ and its graphical representation is $X - Y$.
\end{definition}

\begin{definition}[Directed Graph]
A directed graph (DG) $G$ is a graph where the edge $(X, Y)$ is distinct from the edge $(Y, X)$.
\end{definition}

In particular, a directed edge $(X, Y)$ is graphically represented by an arrow as $X \rightarrow Y$, and induces a set of relationships between the vertices of the graph $G$. Given a vertex $X$, we denote by $Pa(X)$ its {\em parents}, i.e. the set of vertices that have an arrow into $X$, while we denote by $Ch(X)$ its {\em children}, i.e. the set of vertices that have an arrow out of $X$. Recursively, any parent and parent of a parent (child and child of a child) of $X$ is an {\em ancestor} $An(X)$ ({\em descendant} $De(X)$) of $X$.

The vertices connected to $X$ are said to be \emph{adjacent} to $X$ and denoted by $Adj(X)$, while the vertices connected with an undirected edge are the \emph{neighbors} $Ne(X)$. These two sets of vertices are identical in undirected graphs, but may be different in graphs with other mixed orientations.

\begin{definition}[Path]
A path $\pi = (X - \cdots - Y)$ is a tuple of non repeating vertices, where each vertex is connected to the next in the sequence with an undirected edge.
\end{definition}

\begin{definition}[Directed Path]
A directed path $\pi = (X \rightarrow \cdots \rightarrow Y)$ is a tuple of non repeating vertices, where each vertex is connected to the next in the sequence with a directed edge.
\end{definition}

\begin{definition}[Cycle]
A cycle is a path that starts and ends at the same vertex.
\end{definition}

\begin{definition}[Directed Acyclic Graph]
A directed acyclic graph (DAG) is a directed graph $G$ that has no cycles.
\end{definition}

\subsection{Causal Model}

\begin{definition}[Causal Graph]
A \emph{causal graph} $G = (\textbf{V}, \textbf{E})$ \cite{bareinboim20211OP}  is a graphical description of a system in terms of cause-effect relationships, i.e. the \emph{causal mechanism}.
\end{definition}

For instance, the key difference between a Bayesian network (BN) \mbox{\cite{Pearl2018BayesianN}} and a causal Bayesian network (CBN) is the semantic interpretation of the edges. In particular, in a CBN a directed edge $X \rightarrow Y$ establishes a cause-effect relationship between the cause $X$ and its effect $Y$. Hence, reversing an edge in a BN that is not defined by a causal graph might result in the same underlying probability distribution for the pair of variables $X$ and $Y$ due to the Bayes theorem, while reversing an edge in a CBN where the graph is causal means changing the interpretation of the data generating mechanism, i.e. the causal mechanism. This concept is formalized by the following definitions.

\begin{definition}[Direct and Indirect Cause]
For each directed edge $(X, Y) \in \textbf{E}$, $X$ is a \emph{direct cause} of $Y$ and $Y$ is a \emph{direct effect} of $X$. Recursively, every cause of $X$ that is not a direct cause of $Y$, is an \emph{indirect cause} of $Y$.
\end{definition}

This definition is formally enforced by the \emph{causal edge assumption} \cite{glymour2016causal}, as follows:
\begin{definition}[Causal Edge Assumption]\label{eq:causal_edge_assumption}
The value assigned to each $X_i$ is completely determined by the function $f_i$ given its parents:
\begin{equation}
    X_i := f_i(Pa(X_i)) \qquad \forall X_i \in \textbf{V}.
\end{equation}
\end{definition}

As natural consequence of Definition \ref{eq:causal_edge_assumption}, we can define models that entail both the structural representation and the set of functions, i.e. the underlying causal mechanism.

\begin{definition}[Structural Causal Model]
A \emph{structural causal model} (SCM) \cite{glymour2016causal, massmann2021causal} is defined by the tuple $M = (\textbf{V}, \textbf{U}, \textbf{F}, P)$, where:
\begin{itemize}
    \item $\textbf{V}$ is a set of \emph{endogenous} variables, i.e. \emph{observable} variables,
    \item $\textbf{U}$ is a set of \emph{exogenous} variables, i.e. \emph{unobservable}\footnote{\hl{Authors in \mbox{\cite{bareinboim20211OP}} define variables as exogenous if they are determined by factors \textit{outside} the model, for which we choose not to explain the causes, without the precise connotation of unobservable, although typically they are.}} variables, where $\textbf{V} \cap \textbf{U} = \varnothing$,
    \item $\textbf{F}$ is a set of \emph{functions}, where each function $f_i \in \textbf{F}$ is defined as $f_i : (\textbf{V} \cup \textbf{U})^p \rightarrow \textbf{V}$, with $p$ the ariety of $f_i$, so that $f_i$ determines completely the value of $V_i$,
    \item $P$ is a joint probability distribution over the exogenous variables $P(\textbf{U}) = \prod_i P(U_i)$.
\end{itemize}
\end{definition}

Structural Causal Models are also known as Structural Equation Models (SEMs).

\begin{figure}[H]
    \centering
    \begin{minipage}{0.25\linewidth}
        \centering
        \begin{tikzpicture}
            \node (1) at (0, 0) {$X$};
            \node (2) at (0, -1.75) {$Y$};
            \node (3) at (0, -3.5) {$Z$};
            \node[scale=0.75, dashed] (4) at (-1.75, -0.50) {$U_{XY}$};
            \node[scale=0.90, dashed] (5) at (-1.75, -2.25) {$U_Z$};
            \draw[->] (1) to (2);
            \draw[->] (2) to (3);
            \draw[->, dashed] (4) to (1);
            \draw[->, dashed] (4) to (2);
            \draw[->, dashed] (5) to (3);
        \end{tikzpicture}
        \captionsetup{justification=raggedleft, singlelinecheck=false} \subcaption{}
        \label{fig:causal_graph}
    \end{minipage}
    \hspace{1.0cm}
    \begin{minipage}{0.4\linewidth}
        \begin{align*}
            M &= (\textbf{V}, \textbf{U}, \textbf{F}, P) \\
            \textbf{V} &= \{ X, Y, Z \} \\
            \textbf{U} &= \{ U_{XY}, U_Z \} \\
            \textbf{F} &= 
            \begin{cases}
                f_X: X := 2 U_{XY}, \\
                f_Y: Y := X + U_{XY}, \\
                f_Z: Z := 3Y + U_Z
            \end{cases} \\
            P &=
            \begin{cases}
                U_{XY} \sim \mathcal{N}(0,1), \\
                U_Z \sim \mathcal{N}(0,1)
            \end{cases}
        \end{align*}
        \subcaption{}
        \label{fig:scm_definition}
    \end{minipage}
    \caption{The causal graph $G$ (a) of the related SCM $M$ (b). In (a) $X$ is a direct cause of $Y$ and an indirect cause of $Z$, while $Y$ is an effect, and in particular a direct effect, of $X$. An example of associated SCM is reported in (b), where the functional set \textbf{F} follows the causal edge assumption.}
    \label{fig:scm}
\end{figure}
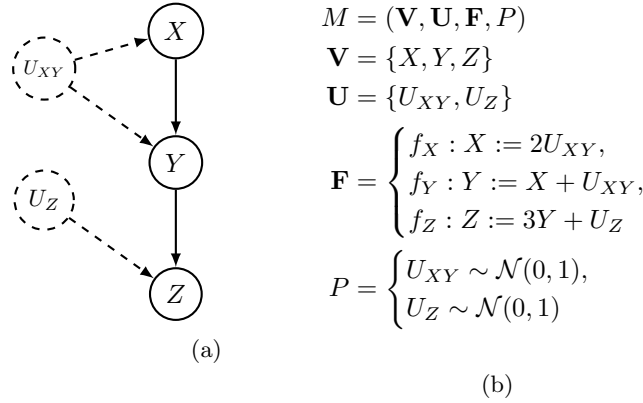

The joint \emph{exogenous distribution} $P$ is responsible for the non-deterministic nature of the model, adding a layer of uncertainty through a set of independent \emph{noise} distributions. The unobserved terms \textbf{U} are represented in Figure \ref{fig:scm} as dashed vertices with dashed edges.

\subsection{The Causal Discovery Problem}
The causal discovery problem \cite{spirtes2016causal} consists in selecting a causal graph as a possible explanation for a given dataset.

Formally, let \textbf{G} be the set of graphs defined over the variables \textbf{V} of a dataset \textbf{D} and $G^* \in \textbf{G}$ be the \emph{true but unknown} graph from which \textbf{D} has been generated.

\begin{definition}[Causal Discovery Problem]
The causal discovery problem \cite{spirtes2000causation} consists in recovering the \emph{true} graph $G^*$ from the given dataset \textbf{D}.
\end{definition}

A causal discovery algorithm is said to \emph{solve} the causal discovery problem if and only if it \emph{converges} to the true graph $G^*$ in the limit of the sample size of the datset \textbf{D}.

\begin{definition}[Soundness and Completeness]\label{def:sound_complete}
A causal discovery algorithm is \emph{sound} if it is able to solve the causal discovery problem, and it is \emph{complete} if it outputs the \emph{most informative} causal graph $G$ that can be recovered from the input dataset \textbf{D}, without making further assumptions.
\end{definition}

While the criterion by which a graph is said to be ``most informative'' depends on the context, in general it refers to the case in which neither the presence of an edge nor its orientation can be modified further without proving more information, i.e. prior knowledge or making additional assumptions.

\begin{definition}[Consistency of a Causal Graph]
A causal discovery algorithm is \emph{consistent} \cite{glymour2019review, spirtes2000causation} if it outputs a graph $G$ that induces a probability distribution consistent with the input dataset \textbf{D}.
\end{definition}

\begin{definition}[Identifiability of a Causal Graph]\label{def:identifiability}
A causal discovery algorithm is said to \emph{identify} \cite{glymour2016causal} a graph $G$ if it is able to determine the direction of any edge in $G$.
\end{definition}

In the following pages we will see that some algorithms are able to identify the causal graph \emph{up-to its equivalence class} (Definition \ref{def:observational_equivalence}), meaning that setting the direction of any of the remaining undirected edges would not induce a different probability distribution, i.e. it is not possible to choose a specific direction for that edge without further assumptions.

Moreover, some of these methods are able to exploit only \emph{observational} distributions, i.e. probability distributions that are induced by observation dataset, while others are capable of taking advantage of \emph{interventional} distributions, i.e. probability distributions that are generated by experimental data, where we intervene on the system of interest.

Finally, even though the general formulation of the causal discovery problem is focused on the causal graph only, causal discovery algorithms are usually designed to find a solution w.r.t. a specific set of functions \cite{bongers2018random, rubenstein2018deterministic, shahbazinia2021paralingam, shimizu2014lingam}, e.g. non-linear equations.

\subsection{Acyclicity and Faithfulness}\label{def:faithfulness}
\begin{definition}[Markov Property]
A graph $G = (\textbf{V}, \textbf{E})$ is said to satisfy the \emph{Markov property} if the associated joint probability distribution $P(\textbf{V})$ can be decomposed \emph{recursively} as:
\begin{equation}\label{eq:factorization}
    P(\textbf{V}) = \prod_{X_i \in \textbf{V}} P(X_i | Pa(X_i))
\end{equation}
\end{definition}

The probability factorization expressed in Equation \ref{eq:factorization} relies on the assumption that the relationships encoded by the graph match exactly the underlying conditional probability independencies:
\begin{equation}\label{eq:conditional_independence}
    X \indep_P Y \, | \, \textbf{Z} \implies X \indep_G Y \, | \, \textbf{Z}
\end{equation}
where \textbf{Z} is a subset of $\textbf{V} \backslash \{X, Y\}$.

Essentially, it is assumed that the probability independence ($\indep_P$) implies the graphical independence ($\indep_G$), as stated in Equation \ref{eq:conditional_independence}.

This assumption is known as \emph{d-faithfulness} or ``directed faithfulness''. In fact, the graphical model is required to rely on a DAG in order to satisfy the Markov property. More recently, extensions of the faithfulness assumption to the cyclic setting have been taken into consideration, e.g. $\sigma$\emph{-faithfulness} \cite{bongers2021foundations, mooij2020constraint}, enabling the discovery of general non-acyclic DGs.

In order to test whether a variable $X$ is conditionally independent from $Y$ given a set \textbf{Z} in any probability distribution $P$ faithful to $G$, one can use the \emph{d-separation} criterion, which is based on the concept of {\em blocked path}.

In particular, when \textbf{Z} \emph{blocks} every path between $X$ and $Y$, we say that $X$ and $Y$ are \emph{d-separated} by \textbf{Z}. A path $\pi$ is blocked depending on the presence of specific graphical patterns in it, as given in the following two definitions.

\begin{definition}[Fork, Chain \& Collider]
Let $G = (\textbf{V}, \textbf{E})$ be a DG and $\pi$ be a path on $G$. Then, given three vertices $X$, $Y$ and $Z$ in $\pi$, we have the following:
\begin{itemize}
    \item $X \leftarrow Y \rightarrow Z$ is a \emph{fork} on $\pi$,
    \item $X \rightarrow Y \rightarrow Z$ is a \emph{chain} on $\pi$, and
    \item $X \rightarrow Y \leftarrow Z$ is a \emph{collider} on $\pi$.
\end{itemize}
\end{definition}

\begin{definition}[d-separation]
Let $G = (\textbf{V}, \textbf{E})$ be a DG, $\pi$ be a path on $G$ and \textbf{Z} a subset of \textbf{V}. The path $\pi$ is \emph{blocked} \cite{glymour2016causal} by \textbf{Z} if and only if $\pi$ contains:
\begin{itemize}
    \item a fork $X \leftarrow Y \rightarrow Z$ or a chain $X \rightarrow Y \rightarrow Z$ such that the middle vertex $Y$ is in \textbf{Z}, or
    \item a collider $X \rightarrow Y \leftarrow Z$ such that middle vertex $Y$, or any descendant of it, is not in \textbf{Z}.
\end{itemize}
The set \textbf{Z} d-separates $X$ from $Y$ if it blocks every path between $X$ and $Y$\footnote{In a more general setting, d-separation can be extended to set of vertices rather than just singletons \cite{pearl1995causal}.}.
\end{definition}

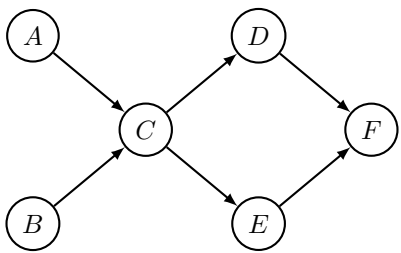
\begin{figure}[h]
    \centering
    \begin{tikzpicture}
        \node (0) at (-1.50, +1.25) {$X$};
        \node (1) at (-1.50, -1.25) {$Y$};
        \node (2) at (+0.00, +0.00) {$Z$};
        \node (3) at (+1.50, +1.25) {$S$};
        \node (4) at (+1.50, -1.25) {$T$};
        \node (5) at (+3.00, +0.00) {$W$};
        
        \draw[->] (0) to (2);
        \draw[->] (1) to (2);
        \draw[->] (2) to (3);
        \draw[->] (2) to (4);
        \draw[->] (3) to (5);
        \draw[->] (4) to (5);
    \end{tikzpicture}
    \caption{In this figure, $X$ and $Y$ are d-separated without conditioning on $Z$, since they form a collider. The same does not hold for $X$ and $S$, given that they form a chain by means of $Z$, and therefore conditioning (i.e. setting its value) on the middle vertex $Z$ d-separates $X$ from $S$.}
    \label{fig:d_separation}
\end{figure}

\subsection{Equivalence Classes}
In the previous paragraphs we introduced the concept of causal graph as natural consequence of the causal edge assumption, where the functional set \textbf{F} is mapped to a directed graph $G$.

The na{\"i}ve representation of a DAG does not allow to convey the (lack of) knowledge that typically arises during a discovery procedure. Here, we define formally other graphical representations, along with their interpretations.

\begin{definition}[Partially DAG]
The graph $G$ is a \emph{partially-directed acyclic graph} (PDAG) if it can contain both undirected $(-)$ and directed $(\rightarrow)$ edges.
\end{definition}

This alternative representation allows to distinguish a cause-effect pair $(X \rightarrow Y)$ from a yet unknown relationship $(X - Y)$, where there is still uncertainty about the direction of the edge.
PDAGs are also called \emph{patterns} \cite{spirtes2000causation}.

\begin{definition}[Skeleton]
Let $G$ be a PDAG. The skeleton of $G$ is the undirected graph resulting from changing any directed edge of $G$ to undirected.
\end{definition}

\begin{definition}[V-structure]
Let $G$ be a PDAG. A v-structure in $G$ is a triple $X \rightarrow Y \leftarrow Z$ where $X$ and $Z$ are not adjacent. V-structures are also called \emph{unshielded colliders} \cite{verma1991equivalence}.
\end{definition}

\begin{figure}[h]
    \centering
    \begin{tikzpicture}
        \node (0) at (+0.00, +0.00) {$A$};
        \node (1) at (-1.25, -1.25) {$B$};
        \node (2) at (+1.25, -1.25) {$C$};
        \node (3) at (+0.00, -2.50) {$D$};
        \node (4) at (+0.00, -4.15) {$E$};
        
        \draw[->] (0) to (1);
        \draw[->] (0) to (2);
        \draw[->] (1) to (3);
        \draw[->] (2) to (3);
        \draw[->] (3) to (4);
    \end{tikzpicture}
    \hspace{0.50cm}
    \begin{tikzpicture}
        \node (0) at (+0.00, +0.00) {$A$};
        \node (1) at (-1.25, -1.25) {$B$};
        \node (2) at (+1.25, -1.25) {$C$};
        \node (3) at (+0.00, -2.50) {$D$};
        \node (4) at (+0.00, -4.15) {$E$};
        
        \draw (0) to (1);
        \draw (0) to (2);
        \draw[->] (1) to (3);
        \draw[->] (2) to (3);
        \draw[->] (3) to (4);
    \end{tikzpicture}
    \caption{A DAG on the left and its CPDAG (Definition \ref{def:cpdag}) on the right. As we can see, both graphs have the same underlying structure (i.e. skeleton), but differ from the orientation of some of the edges. Specifically, the edges connecting $A$ to $B$ and $C$ can be rearranged to form different chains or a fork. This is not true for the others edges in the CPDAG, since they are compelled. In fact, modifying the orientation of one of them would either remove the v-structure formed by $B \rightarrow D \leftarrow C$ or introduce a new one.}
    \label{fig:cpdag}
\end{figure}
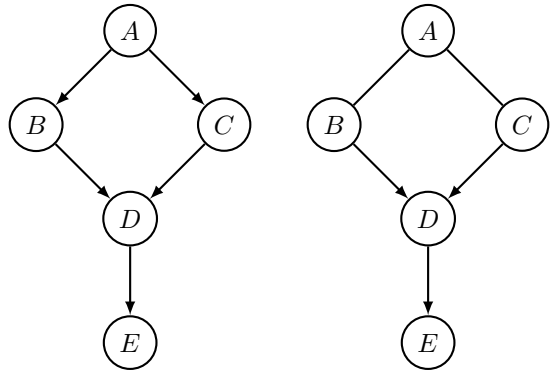

In the context of PDAGs, v-structures encode the conditional independencies that shape the associated probability distribution. Any edge that, when reversed, would either add or remove a v-structure is said to be a \emph{compelled} edge, as in Figure \ref{fig:cpdag}. Any compelled edge, along with the underlying skeleton, is a constraint for the set of observational distributions consistent with the given PDAG. Any non-compelled edge is called \emph{reversible}.

\begin{definition}[Observational Equivalence]\label{def:observational_equivalence}
Two DAGs $G$ and $H$ are \emph{observationally Markov equivalent} \cite{verma1991equivalence} if they have the same skeleton and the same v-structures, denoted as $G \equiv H$.
\end{definition}

Henceforth, the definition of equivalence stems from an observational point of view, where graphs are compared in terms of the observational probability that is faithful to the given structure. In fact, changing the orientation of an reversible edge leads to a different structure with an equivalent factorization of the associated probability distribution.

\begin{definition}[Observational Equivalence Class]
Two DAGs $G$ and $H$ belong to the same \emph{observational Markov equivalence class} (MEC) \cite{mooij2020constraint, mooij2020joint, yang2018characterizing} if they are Markov equivalent. As generalization, the MEC of a graph $G$, denoted by $[G]$, represents the set of possible DAGs that are observationally equivalent.
\end{definition}

Since MECs are defined in terms of skeletons and v-structures only, edges that are not part of any v-structure remain undirected, meaning that, given the limited knowledge, it is not possible to disentangle the relationship between the two variables.

\begin{definition}[Completed PDAG]\label{def:cpdag}
A PDAG $G$ is said to be \emph{completed} \cite{spirtes2000causation} if any directed edge is compelled and any undirected edge is reversible w.r.t. its MEC $[G]$.
\end{definition}

The usual representation of a MEC is a \emph{complete partially-directed acyclic graph} (CPDAG), also called \emph{essential graph} \cite{andersson1997characterization} or \emph{maximally oriented graph} \cite{meek2013causal}. Although the discovery problem is focused on recovering the true graph $G^*$ from a dataset \textbf{D}, it is not always possible to retrieve a \emph{specific} instance, but rather its MEC $[G^*]$.

\subsection{Sufficiency vs. Insufficiency}\label{def:insufficiency}
In many applications, the collected variables are assumed to be sufficient to find the causes of a system of interest. This condition rarely holds true in real world scenarios \cite{kocaoglu2017experimental}.

\begin{definition}[Causally Sufficient Set]
The set of variables \textbf{V} is said to be \emph{causally sufficient} if and only if every cause of any subset of \textbf{V} is contained in \textbf{V} itself.
\end{definition}

That is, there are no \emph{unobserved} variables \textbf{U} that affect the behaviour of the causal mechanism generating the dataset \textbf{D}. If at least one latent cause exists, then \textbf{V} is \emph{causally insufficient}, which means that there exists a non-empty set of unobserved variables \textbf{U} that contains at least a cause of \textbf{V}. In this case, $G$ is only a sub-graph of the \emph{augmented graph} $G^{a}$ \cite{bongers2021foundations, forre2017markov} defined over $\textbf{V} \cup \textbf{U}$, as depicted in Figure \ref{fig:causal_graph}. However, listing which variables are going to be included in \textbf{U} is not an easy task. Unless prior knowledge is available, it is generally assumed \mbox{\cite{glymour2016causal}} that, for each variable $V_i \in \textbf{V}$, there exists one and only one variable $U_i \in \textbf{U}$ that is parent of $V_i$, i.e. there is an edge $U_i \rightarrow V_i$ in $G$. Hence, each endogenous variable will be influenced by one exogenous variable in the simplest scenario, while in complex settings an observed variable may share one or more unobserved parents with others variables in \textbf{V}, i.e. what are usually called \textit{common latent causes}. On the algorithmic side, causal discovery methods, such as the Fast Causal Inference (FCI) \mbox{\cite{zhang2008completeness}}, limit the search space by testing for potential unobserved variables that are common parents of two observed variables, as graphically explained in the figures in the following pages.

The equivalence class related to constraint-based causal insufficient methods relies on the concept of \emph{mixed graph} and its properties.

\begin{definition}[Mixed Graph]\label{def:mixed_graph}
The graph $G$ is a \emph{mixed graph} (MG) \cite{richardson2002ancestral, zhang2008completeness} if it contains undirected $(-)$, directed $(\rightarrow)$ and bidirected $(\leftrightarrow)$ edges.
\end{definition}

In mixed graphs the focus is on the \emph{edge endpoints}, also called \emph{marks}, rather than on the edge itself. For example, the directed edge $X \rightarrow Y$ is decomposed in two marks: the one insisting on $X \!\! \mathrel{{-}{\cdotp}{\cdotp}{\cdotp}}$ and the one insisting on $\mathrel{{\cdotp}{\cdotp}{\cdotp}{\rightarrow}} \!\! Y$. For this reason, we refer to the former as the \emph{tail} mark $(-)$ and the latter as the \emph{arrowhead} mark $(>)$. Therefore, a bidirected edge is an edge with both marks set to arrowheads.

In a bidirected edge $X \leftrightarrow Y$, $X$ is a \emph{spouse} of $Y$ and vice versa. Therefore, the set of vertices connected with a bidirected edge to $X$ is the \emph{spouse set} $Sp(X)$. The graphical relationships inherited from partially directed graphs remain the same.

The fork, chain and collider patterns must be revised in the context of bidirectional edges. Let $G$ be  a MG and $\pi$ a path on $G$. The pattern $X *\!\!\rightarrow Y \leftarrow\!\!* Z$ is a collider on $Y$, where `$*$' stands for a generic mark. Any other pattern is a \emph{non-collider}.

\begin{definition}[M-separation]
Let $G$ be a MG, $\pi$ be a path on $G$ and \textbf{Z} a subset of \textbf{V}. The path $\pi$ is \emph{blocked} \cite{drton2012iterative} by \textbf{Z} if and only if $\pi$ contains:
\begin{itemize}
    \item a non-collider such that the middle vertex is in \textbf{Z}, or
    \item a collider such that middle vertex, or any descendant of it, is not in \textbf{Z}.
\end{itemize}
The set \textbf{Z} m-separates $X$ from $Y$ if it blocks every path between $X$ and $Y$.
\end{definition}

\begin{definition}[Ancestral Graph]
A mixed graph $G$ is \emph{ancestral} if:
\begin{itemize}
    \item $G$ has no (directed) cycles, and
    \item $X \in Sp(Y)$, then $X \not \in An(Y)$, and
    \item $X \in Ne(Y)$, then $Pa(X) = \varnothing \wedge Sp(X) = \varnothing$.
\end{itemize}
\end{definition}

These conditions allow an insightful interpretation of arrowheads in mixed graphs. In particular, in ancestral graphs, an arrowhead implies non-ancestorship, which explains why these representations are particularly useful in defining causal relationships.

\begin{definition}[Maximal Ancestral Graph]
An ancestral graph is \emph{maximal} (MAG) if any pair of non adjacent vertices are \emph{graphically separated} (in terms of m-separation).
\end{definition}

As for the previous definition of the Markov equivalence class of DAGs using a CPDAG, the MEC of a set of MAGs is represented using a \emph{partial ancestral graph} (PAG). A mark that is present in the same location in any MAG of a MEC is called \emph{invariant}.

\begin{definition}[Partial Ancestral Graph]
The graph $G$ is a \emph{partial ancestral graph} (PAG) if it contains any combination of the following edge marks: tail $(-)$, arrowhead $(\rightarrow)$ and circle $(\circ)$. Moreover, let $[G]$ be the MEC associated to $G$, then:
\begin{itemize}
    \item $G$ has the same adjacencies of $[G]$, and
    \item any arrowhead mark in $G$ is invariant in $[G]$, and
    \item any tail mark in $G$ is invariant in $[G]$.
\end{itemize}
\end{definition}

As a direct consequence of this PAG definition, any circle mark present in $G$ represents a variant mark in $[G]$, as for reversible edges of CPDAGs. Thus, PAGs are \emph{the most informative} (Definition \ref{def:sound_complete}) representation of MECs for MAGs, hence, they satisfy the same \emph{completed} definition of CPDAG.

\begin{figure}[h]
    \centering
    \begin{tikzpicture}
        \node (0) at (+0.00, +0.00) {$A$};
        \node (1) at (-1.25, -1.25) {$B$};
        \node (2) at (+0.00, -2.50) {$C$};
        \node (3) at (-1.25, -3.75) {$D$};
        \node (4) at (+1.25, -3.75) {$E$};
        
        \draw[->] (0) to (1);
        \draw[->] (1) to (2);
        \draw[->] (2) to (3);
        \draw (0) to [out=315, in=90,looseness=1] (4);
        \draw[<->] (2) to (4);
    \end{tikzpicture}
    \hspace{0.50cm}
    \begin{tikzpicture}
        \node (0) at (+0.00, +0.00) {$A$};
        \node (1) at (-1.25, -1.25) {$B$};
        \node (2) at (+0.00, -2.50) {$C$};
        \node (3) at (-1.25, -3.75) {$D$};
        \node (4) at (+1.25, -3.75) {$E$};
        
        \draw[o->] (0) to (1);
        \draw[->] (1) to (2);
        \draw[->] (2) to (3);
        \draw[o-o] (0) to [out=315, in=90,looseness=1] (4);
        \draw[<->] (2) to (4);
    \end{tikzpicture}
    \caption{A mixed graph on the left and one of its possible PAGs on the right.}
    \label{fig:pag}
\end{figure}
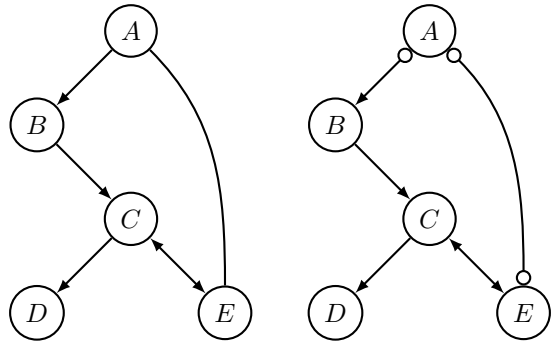

The interpretation of PAGs can be tricky:
\begin{enumerate}
    \item $(X \rightarrow Y)$ : $X$ causes $Y$ and $Y$ does \emph{not} causes $X$, there may be an unobserved confounder,
    \item $(X \leftrightarrow Y)$ : Neither $X$ causes $Y$ nor $Y$ causes $X$, there is an unobserved confounder that causes both $X$ and $Y$,
    \item $(X \circ\!\!\!\rightarrow \! Y)$ : Either $X$ causes $Y$, or there is an unobserved confounder that causes $X$ and $Y$. In this case, $Y$ does \emph{not} causes $X$.
    \item $(X \circ\!\!-\!\!\circ Y)$ : Exactly one of the following holds: $X$ causes $Y$ or vice versa; there is an unobserved confounder that causes $X$ and $Y$; or both (1) and (3) hold; or both (2) and (3) hold.
\end{enumerate}
Understanding which causal statement is implied by each equivalence class is fundamental for a coherent interpretation of its graphical representation, as made explicit in Fig. \ref{fig:pag}.

Depending on additional assumptions, such as homoscedasticity or non-linearity, some algorithms are able to identify the causal graph beyond its equivalence class and recover a single graph instance \cite{peters2017elements, shimizu2014lingam, shimizu2020semiparametric}.

\subsection{Adding Prior Knowledge}
Sometimes a cause-effect pair is known to exist (or to not exist) \emph{a priori}, e.g. through expert's elicitation. Following the causal edge assumption, we can explicitly represent pairs as directed edges, defining a knowledge base composed of required (or forbidden) causal statements.

\begin{definition}[Knowledge Base]
A \emph{knowledge base} $K$ is defined as an ordered pair $(\textbf{R}, \textbf{F})$, where \textbf{R} is the set of required directed edges, while \textbf{F} is the set of forbidden directed edges.
\end{definition}

The knowledge base $K$ is a valid representation for the given \emph{background knowledge}. There exists a class of algorithms that are capable of taking advantage of this prior knowledge \cite{meek2013causal, mooij2020joint}, either by integrating such knowledge before the actual discovery step or by checking if the resulting graph is consistent \emph{a posteriori}.

\begin{table}
    \centering
    \begin{tabular}{|c|c|c|c|c|c|c|c|} \hline
        Algorithm & Year & Category & Output & \rot{Non-Linear} & \rot{Insufficient} & \rot{Cyclic} &  \rot{Intervention} \\ \hline
        PC \cite{colombo2013orderindependent} & 1991 & Constraint & CPDAG & \crossmark & \crossmark & \crossmark & \crossmark \\ \hline
        FCI \cite{zhang2008completeness} & 2008 & Constraint & PAG & \crossmark & \checkmark & \crossmark & \crossmark \\ \hline
        GES \cite{alonso2013scaling} & 2013 & Score & CPDAG & \crossmark & \crossmark & \crossmark & \crossmark \\ \hline
        FGES \cite{ramsey2017million} & 2017 & Score & CPDAG & \crossmark & \crossmark & \crossmark & \crossmark \\ \hline
        ARGES \cite{nandy2018highdimensional} & 2018 & Hybrid & CPDAG & \crossmark & \crossmark & \crossmark & \crossmark \\ \hline
        GFCI \cite{ogarrio2016hybrid} & 2016 & Hybrid & PAG & \crossmark & \checkmark & \crossmark & \crossmark \\ \hline
        HCR \cite{cai2018causal} & 2018 & Score & DAG & \crossmark & \crossmark & \crossmark & \crossmark \\ \hline
        bQCD \cite{tagasovska2020distinguishing} & 2020 & Asymmetric & PDAG & \crossmark & \crossmark & \crossmark & \crossmark \\ \hline
        LiNGAM \cite{hoyer2006estimation, shimizu2014lingam} & 2014 & Asymmetric & DAG & \crossmark & \checkmark & \crossmark & \crossmark \\ \hline
        NOTEARS \cite{zheng2018dags} & 2018 & Score & DAG & \crossmark & \crossmark & \crossmark & \crossmark \\ \hline
        CCD \cite{richardson2013discovery} & 1996 & Constraint & PAG & \crossmark & \crossmark & \checkmark & \crossmark \\ \hline
        LiNG \cite{lacerda2012discovering} & 2012 & Asymmetric & DG & \crossmark & \crossmark & \checkmark & \crossmark \\ \hline
        dseptor \cite{hyttinen2017core} & 2017 & Exact & MG & \crossmark & \checkmark & \checkmark & \crossmark \\ \hline
        bcause \cite{rantanen2020discovering} & 2020 & Exact & MG & \crossmark & \checkmark & \checkmark & \crossmark \\ \hline
        $\sigma$-CG \cite{forre2018constraint} & 2018 & Constraint & $\sigma$-CG & \checkmark & \checkmark & \checkmark & \crossmark \\ \hline
        GIES \cite{hauser2012characterization} & 2012 & Score & CPDAG & \crossmark & \crossmark & \crossmark & \checkmark \\ \hline
        IGSP \cite{yang2018characterizing} & 2018 & Score & CPDAG & \crossmark & \crossmark & \crossmark & \checkmark \\ \hline
        UT-IGSP \cite{squires2020permutationbased} & 2020 & Score & CPDAG & \crossmark & \crossmark & \crossmark & \checkmark \\ \hline
        FCI-JCI \cite{mooij2020joint} & 2020 & Constraint & PAG & \crossmark & \checkmark & \checkmark & \checkmark \\ \hline
        $\Psi$-PC \cite{jaber2020advances} & 2020 & Constraint & CPDAG & \crossmark & \crossmark & \crossmark & \checkmark \\ \hline
        $\Psi$-FCI \cite{jaber2020advances} & 2020 & Constraint & PAG & \crossmark & \checkmark & \crossmark & \checkmark \\ \hline
        backShift \cite{rothenhausler2015backshift} & 2015 & Asymmetric & MG & \crossmark & \checkmark & \checkmark & \checkmark \\ \hline
        bcause+ \cite{rantanen2020learning} & 2020 & Exact & MG & \checkmark & \checkmark & \checkmark & \checkmark \\ \hline
        DCDI \cite{brouillard2020differentiable} & 2020 & Asymmetric & DAG & \checkmark & \crossmark & \crossmark & \checkmark \\ \hline
    \end{tabular}
    \caption{Algorithms classified by supported (\checkmark) and unsupported settings.}
    \label{tab:static_algorithms}
\end{table}

%% file: sections/03_causal_discovery.tex
\section{Causal Discovery}\label{sec:observational}

In this section we introduce the first class of causal discovery algorithms. Here, the hypothetical dataset is represented by static observational data samples, neither interventional information nor time dependencies are taken into account. A summary of the explored algorithms can be found in Table \ref{tab:static_algorithms}.

\subsection{Constraint-based Algorithms}
Constraint-based algorithms try to recover the causal graph by exploiting a set of \emph{conditional independence statements} (CISs) obtained from a sequence of statistical tests. This class of methods translates conditional probability independence into graphical separation by assuming \emph{faithfulness} (Subsection \ref{def:faithfulness}) of the underlying distribution.

\begin{definition}[Perfect Map]
A graph $G$ is said to be a \emph{perfect map} \cite{castillo2012expert, koller2009probabilistic} for a probability distribution $P$ if every CIS derived from $G$ can also be derived from $P$ and vice versa:
\begin{equation}
    X \indep_P Y \, | \, \textbf{Z} \iff X \indep_G Y \, | \, \textbf{Z}
\end{equation}
where \textbf{Z} is a subset of \textbf{V}.
\end{definition}

\begin{definition}[Conditional Independence Test]
The null $H_0$ and alternative hypotheses $H_1$ defined as $H_0: X \indep_P Y \, | \, \textbf{Z}$ and $H_1: X \not \indep_P Y \, | \, \textbf{Z}$, let $I(X, Y | \textbf{Z})$ denote a \emph{conditional independence} (CI) test. The null hypothesis $H_0$ is not rejected if and only if the estimated p-value $\hat{I}(X, Y | \textbf{Z})$ is higher than a chosen significance level $\alpha$:
\begin{equation}
    \hat{I}(X, Y | \textbf{Z}) > \alpha \implies X \indep_P Y \, | \, \textbf{Z}
\end{equation}
where \textbf{Z} is a subset of \textbf{V}.
\end{definition}

When faithfulness is assumed, probability independence implies graphical separation\footnote{Here the term \emph{separation} is used as a placeholder for a generic graphical separation, which is intended as d-separation for directed graphs and m-separation for mixed graphs.}. The main limitation of this approach is related to the exponential growth of the conditioning set \textbf{Z}. Indeed, given the pair $(X, Y)$, in the worst case scenario where $X$ is dependent on $Y$ (or viceversa), the algorithm is required to test for $2^{|\textbf{V} \backslash \{X, Y\}|}$ conditioning sets.

Constraint-based methods are generally capable of integrating prior knowledge into the learning process.

\paragraph{Conditional Independence and Data Types} Constraint-based techniques are essentially \emph{agnostic} of the specific conditional independence test that is being used. Indeed, it is possible to take advantage of such approaches in a wide variety of scenarios, as long as the assumptions of the said test are satisfied. While the main focus of causal discovery studies has been into either discrete or continuous settings, recent advances in conditional independence testing \cite{andrews2019learning, tsagris2018constraint} extend existing tests to mixed-data.

\subsubsection{Peter-Clark (PC)} One of the most studied algorithm that leverages the CISs is the \emph{Peter-Clark} (PC) algorithm \cite{spirtes2000causation} together with its variants \cite{colombo2013orderindependent, thucduy2019fastpc}.

The first step of the procedure consists in defining a complete undirected graph over the variables of the given dataset \textbf{D}. Subsequently, a sequence of CI tests are performed following an heuristic strategy \cite{colombo2013orderindependent}, in order to minimize the number of tests needed. For instance, it is known that the power of CI test decreases when the size of the conditioning set increases \cite{li2020nonparametric}, due to the curse of dimensionality. A common approach consists in selecting an upper limit to the size of the conditioning set, discarding computational-intensive time-wasting tests with low significance levels.

The obtained independence statements are then used to remove the associated edges and to identify the underlying skeleton. Finally, the remaining edges are oriented according to a set of rules \cite{meek2013causal} that leverage the identified v-structures and acyclicity property.

The resulting equivalence class is returned as a CPDAG, where the remaining undirected edges are reversible for the given observational distribution that arises from the data.

\subsubsection{Fast Causal Inference (FCI)} A first extension of the PC algorithm to the causal insufficient setting (Subsection \ref{def:insufficiency}) is represented by the Fast Causal Inference (FCI) algorithm \cite{spirtes2013causal, zhang2008completeness}. Specifically, the FCI algorithm relaxes both the assumption of no latent confounding \cite{hernan2020whatif} and no selection bias \cite{lee2020generalized} in the observational setting, pushing the causal discovery problem a step closer to real-world scenarios. In this context, the authors leverage the definition of \emph{discriminating path} to derive a new set of orientation rules.

\begin{definition}[Discriminating Path]
Let $G$ be an ancestral graph, a path $\pi = (X - \dots - W - Z - Y)$ between $X$ and $Y$ is a discriminating path for $Z$ if i) $\pi$ contains at least three edges, ii) $X$ is not adjacent to $Y$, iii) $Z$ is adjacent to $Y$, and iv) every vertex between $X$ and $Z$ is a collider on $\pi$ and parent of $Y$.
\end{definition}

Discriminating paths are closely related to the separation sets identified by the PC algorithm: if a path $\pi$ between $X$ and $Y$ is discriminating for $Z$, then $Z$ is a collider on $\pi$ iff every set that separates $X$ and $Y$ does not contains $Z$, otherwise it is a non-collider iff every set that separates $X$ and $Y$ contains $Z$.

\subsection{Score-based Algorithms}

Score-based algorithms are usually structured around the maximization of a measure of fitness of a graph $G$ through a space of possible graphs $\mathbb{G}$ for the observed samples $\textbf{D}$, following a defined \textit{scoring criterion} $\mathcal{S}(G, \textbf{D})$ \cite{chickering2002optimal}:
\begin{equation}
    G^{*} = \argmax_{G \in \mathbb{G}} \mathcal{S}(G, \textbf{D})
\end{equation}
In the next few paragraphs, a set of properties for scoring criteria are introduced, before shifting the focus on an optimal two-step procedure for the causal sufficient scenario (Definition \ref{def:insufficiency}).

\begin{definition}[Decomposable Score] A scoring criterion $\mathcal{S}(G, \textbf{D})$ is \textit{decomposable} if it can be defined as a sum of the scores over a vertex and its parents:
\begin{equation}
    \mathcal{S}(G, \textbf{D}) = \sum_{X_i \in \textbf{V}} \mathcal{S}(X_i, Pa(X_i), \textbf{D})
\end{equation}
\end{definition}

As direct consequence of this property, during the discovery procedure, the score computation can be simplified in terms of local differences of the causal graph.

Moreover, the comparison of scores of two DAGs $G$ and $H$ can be handled by taking into account only the vertices that have different parent sets.

\begin{definition}[Equivalent Score] A scoring criterion $\mathcal{S}(G, \textbf{D})$ is \textit{score equivalent} if $\mathcal{S}(G, \textbf{D}) = \mathcal{S}(H, \textbf{D})$, for each pair of graphs $G$ and $H$ in the same equivalence class.
\end{definition}

A graph $G$ is said to \textit{contain} a probability distribution $P$ if there exists an independence model associated with $G$ that represents $P$ exactly, i.e. $G$ is a perfect map of $P$.

\begin{definition}[Consistent Score] \label{def:consistent_score} Let $\textbf{D}$ be a dataset associated with a probability distribution $P$, and let $G$ and $H$ be two graphs. A scoring criterion $\mathcal{S}$ is said to be \textit{consistent} in the limit of the number of samples if and only if:
\begin{itemize}
    \item If only $G$ contains $P$, then $\mathcal{S}(G, \textbf{D}) > \mathcal{S}(H, \textbf{D})$,
    \item If both $G$ and $H$ contain $P$ and the model associated with $H$ has fewer parameters that the one with $G$, then $\mathcal{S}(G, \textbf{D}) < \mathcal{S}(H, \textbf{D})$.
\end{itemize}
\end{definition}

If a scoring criterion is both decomposable and consistent, then it is \textit{locally consistent}.

\begin{definition}[Locally Consistent Score] Let $G$ be a graph and $H$ the graph resulting from addition of the edge $X \rightarrow Y$ to $G$. A scoring criterion $\mathcal{S}(G, \textbf{D})$ is said to be \textit{locally consistent} if and only if:
\begin{itemize}
    \item $X \not \indep_P Y \, | \, Pa(X) \implies \mathcal{S}(H, \textbf{D}) > \mathcal{S}(G, \textbf{D})$, and
    \item $X \indep_P Y \, | \, Pa(X) \implies \mathcal{S}(H, \textbf{D}) < \mathcal{S}(G, \textbf{D})$.
\end{itemize}
\end{definition}
Explicitly, if a scoring criterion is locally consistent then the score:
\begin{itemize}
    \item Increases when any edge that eliminates an independence constraint that does not hold in the generative distribution is added, and
    \item Decreases when any edge that does not eliminate such a constraint is added.
\end{itemize}
This property guarantees that any deletion of an unnecessary edge will produce a higher score value, allowing the definition of an optimal greedy search algorithm.

One of the most commonly used score criterion is the Akaike Information Criterion (AIC) \mbox{\cite{akaike1974criterion}}:
\begin{equation} \label{eq:aic}
    AIC = 2k - 2 \ln{\hat{L}}
\end{equation}
\noindent where $k$ is the number of parameters of the model and $\hat{L}$ is the maximum value of the likelihood for the given model. Models achieving a lower value of AIC are preferred, i.e. they explain better the observed data. Another common scoring criterion is offered by the Bayesian Information Criterion (BIC) \mbox{\cite{schwarz1978estimating}}, also known as the Schwarz Information Criterion:
\begin{equation} \label{eq:bic}
    BIC = k \ln{n} - 2 \ln{\hat{L}}
\end{equation}
\noindent which differs from AIC due to the parameters penalty term that takes into account the number of observations $n$. Others commonly used scoring critera are the Bayesian Dirichlet equivalent uniform (BDeu) \mbox{\cite{geiger1994learning}} and the Bayesian Dirichlet sparse (BDs) \mbox{\cite{scutari2016empirical}}.

\begin{definition}[Optimal Equivalence Class] Let $[G]^*$ be the equivalence class that is a perfect map of the probability distribution $P$ and \textbf{D} the associated dataset. $[G]^*$ is said to be the \textit{optimal} equivalence class if and only if:
\begin{equation}
    \mathcal{S}([G]^*, \textbf{D}) > \mathcal{S}([G], \textbf{D}) \qquad \forall [G] \neq [G]^*
\end{equation}
\end{definition}
\noindent in the limit of the number of samples, \hl{for any consistent scoring criterion $\mathcal{S}$, following Definition {\ref{def:consistent_score}}}.

\subsubsection{Greedy Equivalent Search (GES)}
The Greedy Equivalence Search (GES) \cite{alonso2013scaling, meek1997graphical} is optimal in the limit of the number of samples \cite{chickering2002optimal}. The first step of the algorithm consists in the initialization of the empty graph $G$. The algorithm is composed by two phases: the \emph{forward search} and the \emph{backward search}. In the forward search phase, i) $G$ is modified by repeatedly adding the edge that has the highest difference in score (i.e. delta score), until there is no such edge that increases the score. In the backward search phase, ii) the edge that again achieves the highest delta score is repeatedly removed. The algorithm terminates once it reaches a local maximum during the backward search phase.

This algorithm is designed to work under causal sufficiency. When this assumption no longer holds, the procedure is known to introduce extra edges as a compensation behaviour for the unobserved relationships. For example, when a fork ($X \leftarrow Y \rightarrow Z$) is present and the middle vertex is indeed latent, GES will likely add an edge between the other two observed vertices of the structure, even if such edge is not present in the true graph. Any algorithm that is based on this technique and does not address the issue directly displays such pattern.

\subsubsection{Fast GES (FGES)}
Score-based algorithms are as fast as the computation of the chosen scoring criterion is. Leveraging the properties of the score function, it is possible to minimize the number of computations needed by storing previous intermediate evaluations. Not only this optimization reduces the computation time considerably, but also allows the application of these methods to high-dimensional datasets \cite{andrews2019learning, ramsey2017million}. This ``fast'' variant of GES (FGES) caches partial graph scores, significantly increasing the memory usage, since relevant fragments of the graph may be considered. Moreover, computationally expensive sections of the algorithm can be parallelized, taking advantage of high performance computing (HPC) settings.

\subsection{Hybrid Algorithms}

With the term ``hybrid'' algorithms we refer to the class of methods that combine constraint-based and score-based approaches to mitigate their drawbacks.

\subsubsection{Adaptively Restricted GES (ARGES)}
Consistency of constraint- and score-based algorithms is usually proved in low-dimensional use cases, where the number of samples is orders of magnitude greater than the number of variables. Hybrid approaches generally lacks a formal and rigorous proof of consistency, leading to undefined behaviour. For this reason, an adaptively restricted variant of GES (ARGES) \cite{nandy2018highdimensional} has been developed, targeting specifically the consistency weakness in both low- and high-dimensional spaces.

The novelty of this hybrid version of GES stems from the concept of \emph{admissible edge}. Let $G$ be a CPDAG and $X$ and $Y$ be a pair of non adjacent vertices on it. Adding an edge between $X$ and $Y$ is admissible for the graph $G$ if i) $X$ and $Y$ are adjacent in the (estimated) skeleton of $G$, or ii) there exists a node $Z$ such that $X \rightarrow Z \leftarrow Y$ is a v-structure in $G$.

From the definition of admissible edge, an equal admissible move consists in adding such edge to the graph and obtain a new equivalent CPDAG. This point is sufficient to prove that the resulting forward phase of ARGES is consistent when restricted to admissible moves (i.e. it is an independence map for the given observational probability distribution \cite{chickering2002optimal}).

\subsubsection{Greedy FCI (GFCI)}
Score-based causal discovery algorithms such as GES and FGES are asymptotically correct, but are not designed to work in a causal insufficient scenario (Definition \ref{def:insufficiency}), where unmeasured confounders are present in the true graph. Constraint-based causal search algorithms, such as FCI, are asymptotically correct even with unmeasured confounders, but often perform poorly on small samples. The Greedy Fast Causal Inference (GFCI) \cite{ogarrio2016hybrid} algorithm combines score-based and constraint-based algorithms improving over the previous results while being asymptotically correct under causal insufficiency.

Specifically, the initial skeleton is obtained by un-orienting the CPDAG resulting from the execution of FGES. Then, the orientation rules of FCI are applied, with only a few slight modifications that rely on original FGES output. This approach leads to an improved accuracy over the distinct constraint- and score-based approaches. As a side effect, additional requirements arise from the union of these methods. For example, not only the conditional independence test is required to be consistent by FCI, but also the associated score must be \emph{locally} consistent due to FGES. This constraint reduces the practical applications to settings where indeed such score exists.

\subsection{Other Methods}

\subsubsection{Hidden Compact Representation (HCR)}
Causal discovery methods for discrete and mixed variables have gained renovated interested in the last few years \cite{andrews2019learning, tsagris2018constraint}. Although additive noise models have been widely used in the context of continuous variables, it is difficult to justify their application with categorical data, where the addition operator between the levels of variables is not well defined.

For this reason, authors in \cite{cai2018causal} developed a new low-dimensional embedding for discrete variables, allowing a (hidden) compact representation (HCR) of the discrete states of such variables. The method follows a two-stage procedure: at first, a discrete variable is deterministically mapped into a low-cardinality representation (e.g. binary), which acts as a proxy for the information contained in the original variable; then, a set of samples are drawn for the new proxy variable using a probabilistic mapping. The overall complexity of the model in controlled using the BIC score, balancing between total fitness and size of parameters.

The authors address the problem of identifiability (Definition \ref{def:identifiability}) of the model and prove that, under mild conditions, the causal graph recovered from observational data is identifiable. The method is tested against both synthetic and real-world data, providing reference values for performance metrics. In these experiments, HCR outperforms linear models in terms of accuracy and sensitivity, especially when the additive noise assumption does not hold.

\subsubsection{Bivariate Quantile Causal Discovery (bQCD)}
The bivariate quantile causal discovery (bQCD) \cite{tagasovska2020distinguishing} technique is designed to uncover cause-effect pairs in the bivariate setting. By re-expressing independence statements in light of the minimum description length (MDL) \cite{rissanen1978modeling}, the authors build a discovery procedure by using \emph{quantile scoring}.

Following \cite{janzing2010causal}, let $X$ and $Y$ be two random variables with joint, marginal and conditional distributions denoted by $F_{XY}$, $F_X$ and $F_{X|Y}$, respectively. The key concept here is that a lower complexity follows from a correct causal orientation of the $(X, Y)$ pair, since it is a more informative representation of the associated data.

Hence, the Kolmogorov complexity $K(F)$ is defined as the length of the shortest program $F$ that outputs $F(X)$. Since $K(F)$ measures the information contained in $F$, authors in \cite{stegle2010probabilistic} state that if $X$ causes $Y$, then $K(F_X) + K(F_{Y|X}) \leq K(F_Y) + K(F_{X|Y})$. The problem is that $K(F)$ cannot be computed in practice. Therefore, the authors rely on the MDL principle as a proxy for the Kolmogorov complexity. Such an approximation can be performed by estimating the population quantiles through nonparametric quantile regression. 

The resulting procedure is robust to outliers and can be generalized to a wide range of distributions, although it requires that \emph{all} population quantiles are computable, which could be a limiting factor in real-world applications.

\subsubsection{Linear Non-Gaussian Acyclic Model (LiNGAM)}
In the context of linear causal models, when causal sufficiency holds, the observed variables can be expressed as a linear combination of the noise terms:
\begin{equation}
    \textbf{x} = \textbf{B} \textbf{x} + \textbf{e}
\end{equation}
where \textbf{B} is matrix consisting of the coefficients of the variables in the associated set of equations, \textbf{x} is the vector of variables and \textbf{e} is the vector of noises.

Here, the exogenous distribution is assumed to be made of mutually independent (possibly non-Gaussian) variables. Solving for \textbf{x} reduces to the identification of the matrix \textbf{A} such that:
\begin{equation}
    \textbf{x} = (\textbf{I} - \textbf{B})^{-1}\textbf{e} = \textbf{A}\textbf{e}
\end{equation}
LiNGAM \cite{shimizu2014lingam, shimizu2020semiparametric} relies on Independent Component Analysis (ICA) \cite{comon1994independent} to identify a possible solution for \textbf{A}. In fact, multiple mixing matrices \textbf{A} are feasible solutions for the given joint probability distribution. This technique is essentially focused on discovering asymmetries in the sample distribution to determine the correct causal ordering. Once such ordering has been discovered, the causal graph is built by recovering all and only the edges coherent with the order.

LiNGAM has been extended later for causally insufficient settings \cite{hoyer2006estimation}. Let \textbf{f} be the vector of latent variables and $\boldsymbol{\Lambda}$ the matrix of the connections strength between \textbf{f} and \textbf{x}, then:
\begin{equation}
    \textbf{x} = \textbf{B} \textbf{x} + \boldsymbol{\Lambda} \textbf{f} + \textbf{e}
\end{equation}
The proposed model can be solved with a variant of ICA, called \emph{overcomplete} ICA, which takes into account the presence of unobserved effects.

LiNGAM consistently estimates the connection matrix \textbf{B}. While standard ICA does not scale well in high-dimensional settings, approximated variants of ICA can be used to compute the components with a predefined fix number of iterations with reasonable precision. This leads to an efficient solution in presence of non-Gaussian noise and causally insufficient datasets.

\subsubsection{Continuous Optimization (NOTEARS)}
In the ``DAGs with NO TEARS'' \cite{zheng2018dags} algorithm, the causal discovery problem is reduced to a continuous optimization problem. The acyclicity constraint is expressed as an equality constraint $h(\textbf{W}) = 0$, where $h$ is a smooth differentiable function that measures the ``DAG-ness'' (i.e. a quantification of the acyclicity violations) of a given adjacency matrix \textbf{W}:
\begin{equation}
    h(\textbf{W}) = tr(e^{\textbf{W} \circ \textbf{W}}) - n = 0
\end{equation}
where $tr$, is the trace operator, $\circ$ is the Hadamard product, $e^*$ is the matrix exponential and $n$ the size of \textbf{W}. Moreover, this function has a rather simple associated gradient:
\begin{equation}
    \nabla h(\textbf{W}) = (e^{\textbf{W} \circ \textbf{W}})^T \circ 2\textbf{W}
\end{equation}
Coefficients in \textbf{W} smaller than a fixed threshold $\omega > 0$ are set to zero, rounding the solution with an arbitrary precision. The evaluation of the matrix exponential is $O(n^3)$, i.e. cubic in the number of vertices. Given the low computational complexity, NOTEARS outperforms existing methods when both the in-degree and the sample size are large. However, even if the authors compared their approach to other methods (e.g. PC, FGES, LiNGAM), this algorithm is focused on exploring the DAG space, rather than addressing Definition {\ref{eq:causal_edge_assumption}} directly, i.e. an \textit{acyclic} graph may not necessarily be a \textit{causal} graph.

\subsection{Comparison between Methods}

Now that different classes of methods have been explored, one might be tempted to ask which approach performs better or is faster. Unfortunately, it is difficult to restrict the focus on a specific algorithm due to both theoretical and practical limitations, but still, one could at least try to compare advantages and limitations across classes of algorithms.

On the theoretical side, only a few contributions \mbox{\cite{spirtes2000causation, zheng2018dags}} provide proofs of the asymptotic time complexity of such algorithms, making the comparison impossible due to the lack of references. On the practical side, there exist cases in which statistical criteria enforced during the evaluation steps of different algorithms do not match \mbox{\cite{tsamardinos2006max, niinimaki2012local}}, and others where computational time is recorded using absolute measurements \mbox{\cite{niinimaki2012local, natori2015constraint}}, i.e. seconds. These approaches not only make the comparison across different hardware configurations unfair, but are also affected by the size of the input. Furthermore, these issues are exacerbated by the fragmentation of the software packages across multiple programming languages, which are known to have different degrees of efficiency.

Nonetheless, a recent contribution \mbox{\cite{scutari2019learns}} addresses these experimental pitfalls and conducts a rigorous analysis based on both synthetic and real-world data. While the authors do not exclude the existence of others potentially confounding factors, they essentially conclude that there is no systematic difference across classes of algorithms in terms of accuracy and speed, even taking into account a wide range of scenarios. Other works \mbox{\cite{spirtes2010introduction, koller2009probabilistic}} state that constraint-based algorithms are fast in general, but early mistakes undermine the construction of the final structure due to the chained evaluation of conditional independence tests, especially in high dimensional settings.

These last considerations seem to push towards score-based approaches, also considering the possibility of leveraging parallelization during the score evaluation, but further experiments are needed \mbox{\cite{scutari2019learns}} in order to take into account others methods properly.

%% file: sections/04_with_cycles.tex
\section{Causal Discovery with Cycles}\label{sec:cyclical}

\subsection{Cyclic SCM}
In a SCM, the causal graph induces a functional set \textbf{F} where equations follow the decomposition enforced by the causal edge assumption, Subsection \ref{eq:causal_edge_assumption}. If the causal graph is acyclic, then the SCM itself is called acyclic, or \emph{recursive} SEM. The concept of recursion is linked to the hierarchical order that arises from the topological ordering of the underlying DAG. Indeed, it is possible to define a sequence $X_1, X_2, \dots, X_n$ of vertices over \textbf{V} such that for any $X_i$ and $X_j$ where $i < j$, $X_j$ is not a cause of $X_i$ \cite{berry1984nonrecursive}.

Therefore, in a \emph{non-recursive} SEM, or cyclic SCM, some endogenous variables are connected to each other, forming cycles that do not allow a recursive decomposition. Still, the causal edge assumption is satisfied, since its definition is consistent even in the presence of cycles.

\subsection{No Acyclicity Assumption}
Conditional independencies arising from cyclic SCMs are entailed by the cyclic graphs \cite{richardson2013discovery}. It can be shown that, in general, there is no DAG encoding the conditional independencies that hold in such SCM \cite{nagase2017identifiability}. Nonetheless, cyclic SCMs are widely used to model systems with \emph{feedback}, and are applied in sociology, economics and biology, making this class of models a relevant target of interest for causal discovery techniques.

To test for such independencies, d-separation can be adapted to the cyclic setting under the assumption of causal sufficiency (Definition \ref{def:insufficiency}) \cite{spirtes2013directed}. In causally insufficient scenarios, d-separation can be replaced with $\sigma$-separation \cite{forre2018constraint, forre2017markov} applied to directed mixed graphs (DMGs), i.e. mixed graph (Subsection \ref{def:mixed_graph}) without undirected edges.

\begin{definition}[Strongly Connected Component]
Let $G$ be a DG and $X$ a vertex in $G$. The \emph{strongly connected component} \cite{forre2018constraint} of a vertex $X$ is defined as:
\begin{equation}\label{eq:scc}
    SCC(X) = An(X) \cap De(X)
\end{equation}
that is, the set of vertices that are both ancestors and descendants of $X$, including $X$ itself.
\end{definition}

\begin{definition}[$\sigma$-separation]
Let $G$ be a DMG, $\pi$ be a path on $G$ and \textbf{Z} a subset of \textbf{V}. The path $\pi$ is \emph{blocked} \cite{forre2018constraint, forre2017markov} by \textbf{Z} if and only if $\pi$ contains:
\begin{itemize}
    \item a collider $X *\!\!\rightarrow Y \leftarrow\!\!* Z$ where $Y \not \in An(\textbf{Z})$, or
    \item a non-collider $X \leftarrow Y *\!\!-\!\!* Z$ (or $X *\!\!-\!\!* Y \rightarrow Z$) where $Y \in An(\textbf{Z})$ and $X$ (respectively $Z$) is part of $SCC(Y)$ (Equation \ref{eq:scc}).
\end{itemize}
The set \textbf{Z} $\sigma$-separates $X$ from $Y$ if it blocks every path between $X$ and $Y$.
\end{definition}

The above graphical criterion implies d-separation and reduces to it in the case of DAGs.

\subsubsection{Cyclic Causal Discovery (CCD)}
The Cyclic Causal Discovery (CCD)  algorithm \cite{richardson2013discovery} has been the only provably sound (Subsection \ref{def:sound_complete}) approach to general directed graphs until the development of LiNG \cite{lacerda2012discovering} (Subsection \ref{alg:ling}). CCD is a constraint-based algorithm that follows the same initial procedure as the one of the PC algorithm, with five different orientation rules. CCD outputs a PAG $G$ that differs from the output of FCI for a couple of additional patterns:
\begin{itemize}
    \item underlining triples ($X *\!\!-\!\!* \underline{Y} *\!\!-\!\!* Z$), where $Y$ is an ancestor of \emph{at least} one of $X$ or $Z$ in every graph in $[G]$, and
    \item dotted underlining triples ($X *\!\!\rightarrow\!\! \dotuline{\, Y \,} \!\!\leftarrow\!\!* Z$), where $Y$ is not a descendant of a common child of $X$ and $Z$.
\end{itemize}
These additional patterns arise from a fundamental problem: the algorithms is \emph{not complete} (Definition \ref{def:sound_complete}), and, therefore, there may be features common to all graphs in the same equivalence class that are not present in the output PAG (i.e. it is not the most informative PAG). While not being complete in the same sense of the previous algorithms, CCD is \emph{d-separation complete}, meaning that the resulting PAG represents an equivalence class with a single graph, i.e. it encodes all the needed conditional independencies. Therefore, CCD is useful when one is interested in querying the resulting graph about dependencies, but lacks the capability to represent every causal edge by definition, in contrast to others algorithms. This limitation makes CCD less suitable for the definition of SCMs, especially when one is interested in the form of the functional set.

\subsubsection{Linear Non-Gaussian (LiNG)}\label{alg:ling}
The LiNGAM algorithm can be adapted to the cyclic setting by weakening the acyclicity assumption. Specifically, instead of targeting a DAG, LiNG (or LiNG-D family) \cite{lacerda2012discovering} tries to recover a simple graph (i.e. without self-loops) by forcing all entries on the diagonal of the \textbf{B} matrix to be zero.

While LiNGAM output could be seen as a set of admissible models that contains a single model (i.e. the model is \emph{identifiable}), the cyclic variant usually admits more than one causal graph at a time. In fact, the acyclicity assumption that allowed to find the row-permutation of \textbf{B} that best fits the given dataset is missing. The authors then suggest to limit the discovery procedure to the $k$-th best assignment, following the intuition that permutations associated to inadmissible models would score poorly asymptotically. This approach selects one single model from the equivalent class (i.e. returning set).

LiNG inherits both limits and strengths of the original method: approximate (or sparse) ICA can be a valid alternative if running the full ICA is computationally expensive for the considered task.

\subsubsection{$\sigma$-Connection Graphs}
From the concept of $\sigma$-separation, one can derive a MG where conditional independencies are expressed in the presence of cycles and latent variables, namely a $\sigma$-Connection Graph ($\sigma$-CG). An algorithm to learn this structure from data has been developed \cite{forre2018constraint} as a natural extension of the work presented in \cite{hyttinen2014constraint}. The causal discovery problem is re-casted as a continuous optimization problem based on the following loss function:
\begin{equation}\label{eq:sigma_loss_function}
    \mathcal{L}(G, \textbf{S}) = \sum_{i=1}^{n} \lambda_i (
        \mathbbm{1}_{\lambda_i > 0} -
        \mathbbm{1}_{X_i \indep_G Y_i | \textbf{Z}_i}
    )
\end{equation}
where \textbf{S} is a set of conditional independence statements expressed as $\textbf{S} = \big((X_i, Y_i, \textbf{Z}_i, \lambda_i)\big)^n_{i=1}$, $X_i, Y_i$ and $\textbf{Z}_i$ are variables in \textbf{V}, $\lambda_i \in \mathbb{R} \cup \{ -\infty, +\infty \}$ encodes the confidence of probabilistic conditional independence $X_i \indep_P Y_i | \textbf{Z}_i$ as a constraint and $\mathbbm{1}$ is the indicator function which assumes the value one when the constraint is satisfied.

The $\lambda_i$ weights are evaluated using the indicator function $\mathbbm{1}$ to constrain the conditional dependence between variables. Therefore, Equation \ref{eq:sigma_loss_function} quantifies the observations against the proposed causal graph based on the observed data. During the experimental evaluation, authors relied on weights proposed in \cite{magliacane2017ancestral}:
\begin{equation}
    \lambda_i = \log p_i - \log \alpha
\end{equation}
with $p_i$ representing the p-value of a statistical test for conditional independence and $\alpha$ being a significance level.

Minimizing the loss function may lead to multiple optimal solutions, where each solution $G$ is an instance of the actual equivalence class $[G]$. Indeed, as for d-separation and CPDAGs, the $\sigma$-separation criterion and the associated $\sigma$-CGs take into account possible undirected edges that are invariant for any causal graph belonging to the same equivalence class $[G]$.

This algorithm has been benchmarked against synthetic data in a low-dimensional setting. While the recovery metrics show consistent performance across the experiments, especially when increasing the number of interventions, it is clear that the main limitation of this approach is linked with the $\sigma$-separation encoding, as noted in \cite{rantanen2020learning}. Indeed, the separation checks are performed using Answer Set Programming (ASP), a declarative logic programming language, which slows down the learning procedure.

\subsubsection{bcause}
The procedures described so far are essentially \emph{approximate} algorithms that reduce the search space (i.e. the number of conditionally independence tests) by using previously computed test results. In fact, edges that are tested in later phases rely on adjacent vertices that are selected in earlier steps of the algorithm. During the last few years, \emph{exact} search approaches have been developed in a \emph{branch-and-bound} fashion.

The {\em bcause} algorithm \cite{rantanen2020discovering} explores the search space in a tree-like visit guided by an objective function that determines the \emph{weight} of a potential solution. During the discovery phase, any edge of an intermediate result $G$ is either \emph{absent, present} or \emph{undecided}. Before the actual branching step, the lower bound of the given objective function for the current partial solution $G'$ is computed. If such bound is higher than the weight obtained by the previous solution $G$, the branch can be closed and the algorithm backtracks. Otherwise, if $G'$ contains at least one undecided edge, the procedure branches recursively in two directions: one in which said edge is set as present and the other marked as absent. Finally, if the branch cannot be closed and $G$ has no undecided edges, then the current solution $G'$ is updated if and only if the evaluation of the objective function results in a lower weight. The search procedure will return $G$ as a globally optimal solution.

Since the causal discovery problem is inherently exponential, an exact search algorithm is unfeasible in the general setting. However, if both the objective function and its lower bound can be efficiently evaluated, a constrained space for a low dimensional problem can be effectively explored. For example, the authors benchmark their method under different conditions, showing that assuming acyclicity results in a lower execution time. Moreover, the algorithm maintains a set of constraints satisfied by the local solution and updates them incrementally. Therefore, any incompatible extension of the current solution is ruled out by leveraging a linear programming solver, reducing the total number of evaluations needed.

%% file: sections/05_with_interventions.tex
\section{Causal Discovery with Interventions}\label{sec:interventional}
This section is focused on the difference between learning causal models using either observational or interventional data. While the former setting has been explored extensively in the past decades, only recently solutions for properly handling experimental data have been proposed.

\subsection{Observational vs. Interventional}
In order to grasp the added value of experimental data, we will introduce the concept of \emph{ladder of causality} \cite{bareinboim20211OP, pearl2018why} as a reference framework.

\paragraph{The Ladder of Causation} The ladder of causation, also called the \emph{causal hierarchy}, is an ordered structure composed by three layers, where each layer is mapped to a \emph{cognition} level: observational, interventional and counterfactual. A level inherently defines the set of potential queries that can be answered with the given information associated to it.

In practice, the observational layer is composed by associational or \emph{factual} data, while the interventional layer is related to data that are generated by an \emph{intervention} on the system, i.e. an experiment. Interacting with the system itself is the reason why these two levels are different. The counterfactual layer is the highest level of cognition, where one may ask what would have happened if a different intervention had been performed, opposed to the one that factually altered the system. This hypothetical scenario is strongly opposed to the observational one, being in the \emph{counter-factual} space.

Even if the three layers represent different information levels, they are not distinct. In fact, each layer is a generalization of the previous one, e.g. the observational setting can be seen as a special case of the interventional scenario, where no intervention is performed. Therefore, the interventional layer \emph{subsumes} the observational one. The same happens with the counterfactual layer w.r.t. the interventional one, provided that the former allows to define hypothetical actions that were not present in the latter, as expressed in Table \ref{tab:causal_hierarchy}.

\begin{table*}[t]
    \centering
    \begin{tabular}{|c|p{5cm}|p{2.3cm}|} \hline 
        Layer
        & Question
        & Method
        \\ \hline
        Observational
        & How would \emph{seeing} $X$ change my belief in $Y$?
        & Un/Supervised Learning
        \\ \hline
        Interventional
        & What happens to $Y$ if I \emph{do} $X$?
        & Reinforcement Learning
        \\ \hline
        Counterfactual
        & What would have happened to $Y$ if I \emph{had done} $X'$ instead of $X$?
        & Structural Causal Model
        \\ \hline
    \end{tabular}
    \caption{Layers of causation with associated questions, practical examples and methods.}
    \label{tab:causal_hierarchy}
\end{table*}

At this point, one may ask how to formally represent the concepts expressed by this hierarchy, to operatively exploit the informative gap between the layers. The answer is provided by \emph{$do$-calculus} \cite{pearl1995causal}.

\paragraph{$do$-calculus} Queries that are usually expressed in natural language can be rephrased in terms of probability distribution by introducing the $do$ operator, whenever possible\footnote{We restrict ourselves to a minimal introduction of the $do$-calculus, aiming to formally represent the set of concepts that are essential for the causal discovery scenario. For a broader discussion on \emph{identification} and \emph{estimation} of the causal effect, refer to \cite{shpitser2008complete}.}. The $do$ operator represents an intervention on a given variable, e.g. $do(X = x)$ sets the value of $X$ to $x$. This notation is usually overloaded by extending the operator over sets of variables, e.g. $do(\textbf{X} = \textbf{x})$ with $\textbf{x}$ a vector of values. Finally, the $\textbf{x}$ vector can be omitted entirely for brevity.

\begin{definition}[Rules of $do$-calculus]
Let $G$ be a causal graph and $P$ the probability distribution induced by $G$. For any disjoint subset of variables \textbf{X}, \textbf{Y}, \textbf{Z} and \textbf{W}, the following three rules apply:
\begin{enumerate}
    \item Insertion and deletion of observations:
    \begin{equation}
        P(\textbf{Y} \, | \, do(\textbf{X}), \textbf{Z}, \textbf{W}) = P(\textbf{Y} \, | \, do(\textbf{X}), \textbf{W})
    \end{equation}
    if $(\textbf{Y} \indep \textbf{Z} \, | \, \textbf{X}, \textbf{W})$ holds true in $G_{\overline{\textbf{X}}}$,
    \item Exchange of observations and interventions:
    \begin{equation}
        P(\textbf{Y} \, | \, do(\textbf{X}), do(\textbf{Z}), \textbf{W}) = P(\textbf{Y} \, | \, do(\textbf{X}), \textbf{Z}, \textbf{W})
    \end{equation}
    if $(\textbf{Y} \indep \textbf{Z} \, | \, \textbf{X}, \textbf{W})$ holds true in $G_{\overline{\textbf{X}},\underline{\textbf{Z}}}$,
    \item Insertion and deletion of interventions:
    \begin{equation}
        P(\textbf{Y} \, | \, do(\textbf{X}), do(\textbf{Z}), \textbf{W}) = P(\textbf{Y} \, | \, do(\textbf{X}), \textbf{W})
    \end{equation}
    if $(\textbf{Y} \indep \textbf{Z} \, | \, \textbf{X}, \textbf{W})$ holds true in $G_{\overline{\textbf{X},\textbf{Z}(\textbf{W})}}$,
\end{enumerate}
where $G_{\overline{\textbf{X}}}$ is the subgraph of $G$ where the incoming edges into \textbf{X} are removed, $G_{\underline{\textbf{Z}}}$ is the analogous for the outgoing edges from \textbf{Z}, and finally $\textbf{Z}(\textbf{W})$ is $\textbf{Z} \, \backslash \, An(\textbf{W})$ w.r.t. the subgraph $G_{\overline{\textbf{X}}}$.
\end{definition}

With these rules, which are \emph{correct and complete}, a causal effect can be identified if there exists a finite sequence of applications of such rules leading to a $do$-free expression of the considered probability distribution.

\subsection{Types of Interventions}
\begin{definition}[Perfect Intervention]
An intervention is said to be \emph{perfect} (or \emph{hard}) if it removes the causal dependencies (i.e. the incoming causal edges, as in Subsection \ref{eq:causal_edge_assumption}) that affect the intervention target.
\end{definition}

Indeed, $do$-calculus enables us to express perfect interventions in an operative framework, but there are other types of interventions that cannot be expressed using this notation.

\begin{definition}[Imperfect Intervention]
An intervention is said to be \emph{imperfect} (or \emph{parametric}, \emph{soft}) \cite{markowetz2005probabilistic} if it does not remove the causal dependence that affects the intervention target, but alters the functions that represent such dependence.
\end{definition}

For instance, an imperfect intervention on an SCM could be a change in the parameters that quantify the strength of the causal relationships, while a perfect intervention would result in hard setting them to zero. In this sense, perfect interventions are a subset of imperfect interventions, where some variables are removed from the equations of the functional set as a special case.

\paragraph{Mechanism Change} Imperfect interventions itself are a formal definition of a broader concept called \emph{mechanism change} \cite{tian2013causal}. For a SCM $M$ with a causal graph $G$ and a set of parameters $\Theta$ associated to the function set \textbf{F}. A mechanism change is a mapping from $M$ to $M'$, where the new set of parameters is defined as $\Theta' = \Psi' \cup (\Theta \, \backslash \, \Psi)$, with the new subset $\Psi'$ that differs from the original subset $\Psi$. The change affects the behaviour of the function set \textbf{F}, inducing a set $\textbf{F}'$.

\subsection{Defining the Intervention Target}
We can rephrase perfect and imperfect interventions under a single unified framework through the concept of intervention target \cite{hauser2012characterization}.

\begin{definition}[Intervention Target]
Let $G$ be a causal graph. A subset $\textbf{I} \subset \textbf{V}$ is said to be an \emph{intervention target} if it contains all and only the variables associated to an \emph{intervention} over $G$.
\end{definition}

Therefore, a single-variable intervention is an intervention target that contains only one variable, while in a multi-variable intervention it contains more than one. As a special case, when $\textbf{I} = \varnothing$ the intervention target represents the observational case. A set of multiple intervention targets $\{ \textbf{I}_0, \textbf{I}_1, \dots, \textbf{I}_n \}$ is called an \emph{intervention family} and it is denoted with the calligraphic letter $\mathcal{I}$.

\begin{definition}[Conservative Family]
A family of targets $\mathcal{I}$ is \emph{conservative} if for each vertex $X$ in \textbf{V} there exists at least one intervention target in $\mathcal{I}$ that does not contain $X$:
\begin{equation}
    \exists \textbf{I} \in \mathcal{I}: X \not \in \textbf{I}, \quad \forall X \in \textbf{V}
\end{equation}
\end{definition}
Essentially, a conservative family is a family that allows the existence of at least one intervention target that does not intervene on a specific variable. This property guarantees that there is at least one experiment in the family that does not alter the behaviour of such variable if performed.

In this setting, a conservative family allows to observe the influence of a (known) set of targets on at least one unaffected variable, enabling the possibility of disentangling such effect, especially when compared to the other experiments in the whole family.

\begin{definition}[Intervention Graph]
Let $G$ be a causal graph and \textbf{I} be an intervention target defined over $G$. The \emph{intervention graph} $G^{(\textbf{I})} = (\textbf{V}, \textbf{E}^{(\textbf{I})})$ is the causal graph obtained by removing any directed edge that points to a vertex in \textbf{I} from $G$:
\begin{equation}
    \textbf{E}^{(\textbf{I})} = \{ (X, Y) \; | \; (X, Y) \in \textbf{E} \; \wedge \; Y \not \in \textbf{I} \}
\end{equation}
\end{definition}
This definition of intervention graph is coherent with the intervened graph resulting from a $do$-intervention \cite{pearl1995causal}, also known as \emph{graph surgery} or \emph{graph manipulation}.

We can now formally express the interventional distribution associated to an intervention graph.

\begin{definition}[Interventional Distribution]
Let $G$ be a causal graph and \textbf{I} be an intervention target. The \emph{interventional distribution} $P^{(\textbf{I})}$ can be expressed using the factorization formula:
\begin{equation}
    P^{(\textbf{I})} = \prod_{X_i \in \textbf{I}} P^{(\textbf{I})}\big( X_i | Pa(X_i) \big) \prod_{X_i \not \in \textbf{I}} P^{(\varnothing)}\big( X_i | Pa(X_i) \big)
\end{equation}
where $P^{(\varnothing)}$ is the observational distribution of the variables that were not included in the intervention target, if any.
\end{definition}
In case of perfect interventions, the interventional distribution can also be expressed using the $do$-notation:
\begin{equation}
    P^{(\textbf{I})} = \prod_{X_i \in \textbf{I}} P^{(\textbf{I})}\big( X_i \, | \, do(\textbf{I}) \big) \prod_{X_i \not \in \textbf{I}} P^{(\varnothing)}\big( X_i | Pa(X_i) \big)
\end{equation}

\begin{definition}[Interventional Equivalence]\label{def:i_equivalent}
Let $G$ and $H$ be two causal graphs and $\mathcal{I}$ be an intervention family. $G$ and $H$ are \emph{interventionally Markov equivalent} w.r.t. the family $\mathcal{I}$ (i.e. $\mathcal{I}$-equivalent) if the associated intervention graphs $G^{(\textbf{I})}$ and $H^{(\textbf{I})}$ have the same skeleton and the same v-structures for each intervention target of the family:
\begin{equation}
    G \equiv_\mathcal{I} H \implies G^{(\textbf{I})} \equiv H^{(\textbf{I})}, \quad \forall \textbf{I} \in \mathcal{I}
\end{equation}
\end{definition}
In other terms, interventional equivalence can be decomposed in a set of equivalence statements of intervention graphs, where each observational equivalence statement is formulated against a single intervention target contained in the given family.

\begin{definition}[Interventional Equivalence Class]
Two causal graphs $G$ and $H$ belong to the same \emph{interventional Markov equivalence class} w.r.t. the intervention family $\mathcal{I}$ ($\mathcal{I}$-MEC) \cite{jaber2020advances, kocaoglu2019advances} if they are $\mathcal{I}$-equivalent. As for the observational setting, the $\mathcal{I}$-MEC of a graph $G$, denoted by $[G]_\mathcal{I}$, represents the set of possible causal graphs that are interventionally equivalent.
\end{definition}

An intervention family $\mathcal{I}$ induces a classification of the edges of an intervention graphs depending on the effect on the underlying interventional distribution.

\begin{definition}[$\mathcal{I}$-covered edge]
An edge $(X \rightarrow Y)$ in $G$ is $\mathcal{I}$\emph{-covered} if:
\begin{equation*}
    Pa(X) = Pa(Y) \, \backslash \, \{X\} \, \wedge \, P^{(\{X\})}(Y) = P^{(\varnothing)}(Y)
\end{equation*}
when the intervention target $\{X\}$ is in $\mathcal{I}$.
\end{definition}

\begin{definition}[$\mathcal{I}$-contradictory edge]
An edge $(X \rightarrow Y)$ in $G$ is $\mathcal{I}$\emph{-contradictory} if at least one of the following conditions holds:
\begin{itemize}
    \item $\exists \textbf{S} \subset Ne(Y) \, \backslash \, \{X\}$ such that $\forall \textbf{I} \in \mathcal{I}_{X \backslash Y}$ we observe $P^{(\textbf{I})}(Y \, | \, \textbf{S}) = P^{(\varnothing)}(Y \, | \, \textbf{S})$, or
    \item $\forall \textbf{S} \subset Ne(X) \, \backslash \, \{Y\}$ such that $\exists \textbf{I} \in \mathcal{I}_{Y \backslash X}$ we observe $P^{(\textbf{I})}(X \, | \, \textbf{S}) \neq P^{(\varnothing)}(X \, | \, \textbf{S})$.
\end{itemize}
\end{definition}

$\mathcal{I}$-contradictory edges are particularly of interest since they differs among interventional equivalence classes, i.e. they violate the $\mathcal{I}$-Markov property, highlighting the possibility for a consistent exploitation during the discovery procedure.

\subsection{Learning with Interventions}
Sometimes researchers want to observe the effect of an intervention on one single variable at a time, but there are settings in which this is not possible or it is inconvenient. Therefore, multi-variable interventions must be addressed as a special case of a generic intervention target.

\paragraph{Single vs. Multi-Variable Interventions}
When each intervention target contains a single variable at a time, the number of experiments needed to collect enough evidence to identify the causal graph is $n-1$, with $n$ the number of variables \cite{eberhardt2012number}. Indeed, if one intervention would enable the identification of the causal edges incoming into the only variable contained in the intervention target, then the $n$-th intervention would be redundant.

In the case of intervention targets with more than one variable, only $\floor{\log(n)} + 1$ interventions\footnote{Where $\floor{x}$ denotes the \emph{floor} function that maps a real number $x$ to the greatest integer less than or equal to $x$.} are necessary and sufficient in the worst case scenario \cite{eberhardt2012number}, where the causal graph is the complete graph. Since this worst case is improbable, $O(\log\log(n))$ can be achieved as lower bound with high probability in the multi-variable setting with a randomized intervention scheme \cite{hu2014randomized}, that is, it is possible to plan the experimental design in advance to minimize the number of interventions.

\paragraph{Unknown Intervention Targets}
An other problem that one may face during structural learning with interventional data is the uncertainty related to the interventional targets \mbox{\cite{squires2020permutationbased, jaber2020advances, rothenhausler2015backshift}}. There are scenarios in which it is known that an intervention has been performed, but it is unclear which is the exact set of variables that has been affected by such intervention. In this case, an additional layer of complexity is added in order to properly handle the less informative setting of \emph{unknown intervention targets}.

\subsection{Interventional Algorithms}

\subsubsection{Interventional GES (GIES)}
By leveraging the similarity between observational causal graphs and their interventional counterparts, authors in \cite{hauser2012characterization} proposed a generalization of the GES algorithm to the interventional setting. This new score-based variant, called Greedy Interventional Equivalence Search (GIES), follows the same two step approach of the original procedure, traversing the search space using forward- and backward- phases, until a (local) maximum score is reached.

A major contribution of this work is related to the formalization of the interventional setting. Indeed, while the algorithm itself does not differ significantly from the observational one in terms of overall design, the performance improvements are relevant, as expected by transitioning from the first to the second layer of the causal hierarchy. This is an interesting example of how observational techniques can be adapted to the interventional setting with ease, once the theoretical aspects of both the intervention distribution and the intervention targets are addressed properly.

\subsubsection{Interventional Greedy Permutation (IGSP)} While GIES focuses its attention on perfect interventions, a first extension to general interventions is presented in \cite{yang2018characterizing}, with the \emph{Interventional Greedy Sparsest Permutations} (IGSP), an interventional variant of the GSP \cite{solus2021consistency}. In this case, the \emph{greedy} approach consists in the optimization of a score function, coupled with a permutation-based strategy that guides the traversal of the $\mathcal{I}$-MECs space.

Formally, let $\rho$ be a permutation of vertices of a causal graph $G$. The space on which such permutation lays is a polytope called \emph{permutahedron}. A possible representation of this mathematical object is indeed another graph, where each vertex corresponds to a permutation $\rho$ and each edge between two permutations encodes a transposition of the vertices. The goal of a permutation-based causal discovery algorithm is to find a permutation $\rho^*$, consistent with the topological order of the true causal graph $G^*$, that optimizes a given score function. The search procedure traverses the permutahedron using a depth-first approach starting from an initial permutation $\rho$. For each permutation $\tau$ visited, if $G_\tau$ yields a better score than $G_\rho$ then $\rho$ is set to $\tau$. The traversal is restarted from the updated $\rho$, until no such $\tau$ is found.

In order to leverage the advantages of the interventional data, IGSP limits the vertices transposition to the neighbors that are connected by $\mathcal{I}$-covered edges, restricting the search space to permutations that are coherent with the intervention targets. An other characteristic of this search strategy is given by the prioritization of $\mathcal{I}$-covered edges that are also $\mathcal{I}$-contradictory, given that they represent a transition of $\mathcal{I}$-MEC, which could lead to an improvement of the total score.

An extended version of this algorithm, named \emph{UT-IGSP}, has been presented in \cite{squires2020permutationbased} in order to tackle the \emph{unknown target} scenario. The main contribution of this work is linked to the new definition of $\mathcal{I}$-covered edges in light of partially unknown intervention targets.

IGSP (and later UT-IGSP) has been compared to GIES under different conditions, showing that the former achieves better performances than the latter when the dimensionality of the problem is limited (i.e. lower than 10 vertices). This limit is coherent with others traversal-based approaches: although GIES is not consistent in general, its score function is more efficient in pooling together the various interventional datasets when it comes to high-dimensionality spaces.

\subsubsection{Joint Causal Inference with FCI (FCI-JCI)}
Another formal approach, similar to the one introduced in the previous subsection, is presented under the name of \emph{Joint Causal Inference} (JCI) \cite{mooij2020joint}. This method aims to pool together multiple observations collected during different experiments (i.e. \emph{contexts}), hence, the name \emph{joint} causal inference.

In this framework, the set of observed variables is split into two disjoint sets: \emph{system variables} \textbf{X} and \emph{context variables} \textbf{C}. While the former set contains the variables that have been observed during an experiment, the latter set describes under which conditions such system has been observed, following the classical distinction between endogenous and exogenous variables, respectively.

Context variables can be used as \emph{intervention variables}, even if this might not always be the case: here the term is related to the notion of \emph{change of context}, which is a broader scope than simply intervene on the system. Doing so, it is possible to obtain a more flexible representation of the system of interest, where external forces are represented as internal characteristics of a \emph{meta-system}. This approach relaxes the boundary between experiments performed under different conditions, allowing researchers to join data with a coherent causal description.

Before diving into JCI itself, there are a couple of assumptions that can be (optionally) taken into consideration to understand the purpose of the entire context framework:
\begin{enumerate}
    \setcounter{enumi}{-1} 
    \item The underlying mechanism that generates the data is represented by a SCM $M$, where the observed variables are split in system variables and context variables.
    \item No system variable is cause of any context variable, i.e. \emph{exogeneity assumption}.
    \item No system variable is confounded with any context variable, i.e. \emph{randomized context}.
    \item Let $G_\textbf{C}$ be the \emph{context} graph induced by the context variables \textbf{C} over the causal graph $G$ associated with the SCM $M$. For each pair of context variables $(C_i, C_j)$ in the context graph the following holds true:
    \begin{equation}
        (C_i \leftrightarrow C_j) \in G_\textbf{C} \wedge (C_i \rightarrow C_j) \not \in G_\textbf{C}
    \end{equation}
    that is, no context variable is a direct cause of another context variable, but there is a hidden confounder between each pair of context variables, i.e. \emph{generic context}.
\end{enumerate}
While assumptions (0), (1) and (2) are usually considered mild in the interventional setting, assumption (3) might need to be clarified further: if the goal of the causal discovery is to disentangle the causal relationships \emph{using} the context variables as guidance, rather than \emph{focusing on} the connections of the context graph, then assumption (3) can be enforced if (1) and (2) were also assumed. This approach allows the algorithm to restrict the search space to the graphs that satisfy this last assumption, speeding-up the learning process.

The generic JCI procedure can be \emph{adapted to any} observational causal discovery algorithm by following four steps: i) add the context variables, ii) pool data together by setting the values of the context variables, iii) address faithfulness violations between contexts, if any, iv) execute the selected observational learning algorithm. Authors provide reference adaptations for multiple algorithms, such as FCI.

The FCI-JCI variant is particularly of interest, provided that it inherits the strong points of FCI in the causally insufficient setting. Various combinations of the three assumptions were tested, showing that FCI123 (i.e. all three assumptions made) is less accurate in general, but significantly faster than others solutions, allowing its application in more complex scenarios with a sensible number of variables.

\subsubsection{Unknown Intervention Targets using $\Psi$-FCI}
Authors in \cite{jaber2020advances} adapted both PC and FCI algorithms to the causal discovery setting under imperfect interventions with unknown intervention targets. The fundamental contribution of this work is the extension of the $\mathcal{I}$-MEC to a more general $\Psi$-MEC that is capable of representing intervention graphs with unknown intervention targets.

The key idea is that a pair of intervention targets $\textbf{I}, \textbf{J} \in \mathcal{I}$ can be used to identify a unique interventional mechanism that encompasses both targets. Let $G$ be a causal graph and $\mathcal{I}$ an intervention family. The induced set of interventional probability distributions $P^{(\mathcal{I})} = \{ P^{(\textbf{I}_0)}, P^{(\textbf{I}_1)}, \dots, P^{(\textbf{I}_n)}\}$ satisfies the $\Psi$-Markov property if the following holds true for any \textbf{Y}, \textbf{Z} and \textbf{W} disjoint subsets of variables of \textbf{V}:
\begin{enumerate}
    \item Insertion and deletion of observations:
    \begin{equation}\label{eq:psi_1}
        P^{(\textbf{I})}(\textbf{Y} \, | \, \textbf{Z}, \textbf{W}) = P^{(\textbf{I})}(\textbf{Y} \, | \, \textbf{W})
    \end{equation}
    if $(\textbf{Y} \indep \textbf{Z} \, | \, \textbf{W})$ holds true in $G$ for all $\textbf{I} \in \mathcal{I}$,
    \item Invariance of interventions:
    \begin{equation}\label{eq:psi_2}
        P^{(\textbf{I})}(\textbf{Y} \, | \, \textbf{W}) = P^{(\textbf{J})}(\textbf{Y} \, | \, \textbf{W})
    \end{equation}
    if $\big( \textbf{Y} \indep \textbf{K} \, | \, (\textbf{W} \, \backslash \, \textbf{W}_\textbf{K}) \big)$ holds true in $G_{\underline{\textbf{W}_\textbf{K}}\overline{\textbf{R}(\textbf{W})}}$ for all $\textbf{I}, \textbf{J} \in \mathcal{I}$, where \textbf{K} is the symmetric difference of \textbf{I} and \textbf{J}, $\textbf{W}_\textbf{K} = \textbf{W} \cap \textbf{K}$, $\textbf{R} = \textbf{K} \, \backslash \, \textbf{W}_\textbf{K}$ and $\textbf{R}(\textbf{W}) = \textbf{R} \, \backslash \, An(\textbf{W})$  w.r.t. $G$.
\end{enumerate}
While Equation \ref{eq:psi_1} is essentially derived from observational Markov equivalence, Equation \ref{eq:psi_2} is related to the distributional invariances across pairs of intervention targets w.r.t. the associated intervention graph. Indeed, if \textbf{I} and \textbf{J} are the \emph{true} intervention targets for $P^{(\textbf{I})}$ and $P^{(\textbf{J})}$, they must satisfy the invariance for the interventional distributions when separation holds in a given intervention graph.

Moreover, the $\Psi$-Markov property does not require any assumption about the experimental setting in which such interventions are performed. Specifically, it could happen that a subset of experiments were not carried out exactly in the same way, e.g. not in a \emph{controlled} environment. Therefore, even if interventions targets were known a priori, $\Psi$-Markov would still be more general than the $\mathcal{I}$-Markov property.

The authors then recast the augmented graph proposed by \cite{bongers2021foundations, forre2017markov}, adding a set of \emph{utility} vertices that are analogous to the context vertices proposed by \cite{mooij2020joint}. Therefore, the output of the former can be compared to the latter using the related augmented graph, showing that the accuracy of the edge orientations recovered by their $\Psi$-FCI variant is superior than the one proposed by FCI-JCI.

\subsubsection{backShift}
Continuing in the unknown targets setting, the \emph{backShift} \cite{rothenhausler2015backshift} algorithm is a causal discovery approach that recovers linear (possibly cyclical) models under causal insufficiency. It focuses on \emph{shift interventions} with unknown targets, a subset of imperfect interventions where the effect of such perturbation yields a fixed shift of the intervened variable. Both the targets and the shift value can be estimated from the data.

The key idea of this technique is to represent the target SCM $M$ as:
\begin{equation}
    (\textbf{I} - \textbf{B})\textbf{x} = \textbf{c} + \textbf{e}
\end{equation}
where \textbf{x} is a random vector, \textbf{B} is the adjacency matrix of the casual graph associated to $M$, \textbf{e} is the noise vector and \textbf{c} is the random shift vector that models the shift intervention on the system. Then, a joint matrix diagonalization is applied to the differences between the covariance matrices $\Delta \pmb{\Sigma}$ of each experiment $\textbf{I} \in \mathcal{I}$:
\begin{equation}
    \widetilde{\textbf{D}} = \argmin_{\textbf{D} \in \mathcal{D}} \sum_{\textbf{I} \in \mathcal{I}} \mathcal{L}(\textbf{D} \Delta \pmb{\Sigma}^{(\textbf{I})} \textbf{D}^T)
\end{equation}
where $\textbf{D} = \textbf{I} - \textbf{B}$, $\mathcal{L}$ is the sum-of-squared loss function, $\mathcal{I}$ the family of targets.

This approach assumes that data represent observation at the equilibrium, the \textbf{D} matrix is invertible and the \emph{cycle product} \cite{rothenhausler2015backshift} is strictly smaller than one. Moreover, noises, interventions between variables and between experiments are assumed to be uncorrelated.

Authors compare their solution to the LiNG observational alternative, taking advantage of the interventional asymmetries arising from the additional information contained in the data. The results show that \emph{backShift} is capable of dealing with both interventions and latent variables under mild assumptions, outperforming LiNG in both the observational and interventional settings. Moreover, the computational complexity is $O(|\mathcal{I}| \cdot n^2 \cdot m)$, with $n$ representing the number of the variables and $m$ representing the sample size, which allows its application in high-dimensional settings.

\subsubsection{bcause+}
An extension of the \emph{bcause} algorithm to interventional data, called \emph{bcause+}, is proposed in \cite{rantanen2020learning}. When multiple experimental datasets are available, the core-base estimation of the lower bound of each branch of the exact search can be improved by taking into account the variables affected by the intervention.

In particular, the graphical separation statements checks by the observational variant (using either d-separation or $\sigma$-separation) are extended to consider the constraints induced by an intervention target. By assuming the absence of edges oriented into vertices that are part of an intervention target, the search procedure can avoid to check for separation, e.g. in case of perfect interventions. In this sense, intervention targets can be used to derive linear programming constraints by considering the subsets of intervened variables that affect the separation statements.

The improved version of the previous algorithm is also evaluated on non-linear cyclic causal models, showing its capability to deal with non-linear relationships. However, even with the added constraints, the exponentially-increasing execution time prohibits its application in high-dimensional contexts, which is a well known limitation for exact search methods.

\subsubsection{Differentiable Causal Discovery with Interventions (DCDI)}
Under regularity assumption, authors in \cite{brouillard2020differentiable} propose a general differentiable causal discovery algorithm that is capable of learning causal graphs from interventional data with both perfect and imperfect interventional targets, even in the case of unknown interventions.

The key idea of this algorithm is to maximize a score function defined as follows:
\begin{equation}
    \mathcal{S}_\mathcal{I}(G) = \sup_\phi \sum_{\textbf{I} \in \mathcal{I}} \mathbb{E}_X \log f^{(\textbf{I})}(X; \textbf{B}, \textbf{R}, \phi) - \lambda |G|
\end{equation}
where $\phi$ are the weights of the estimator used to maximize the score function (i.e. neural networks in this case), $X$ follows the interventional distribution $P^{(\textbf{I})}$, $f^{(\textbf{I})}$ is the interventional density function, \textbf{B} the binary adjacency matrix of $G$, \textbf{R} the binary interventional matrix (i.e. $R_{ij} = 1$ if $X_i \in \textbf{I}_j$) and $\lambda$ a penalty coefficient.

Essentially, the score function is built upon the conditional interventional distribution to recover the invariant edges across interventions. In fact, vertices that are not in any intervention target are characterized by a conditional probability distribution that is invariant across interventions, as for conservative families of interventions. Relying on conditional invariance, the causal graph $\hat{G} = \argmax_{G \in \mathbb{G}} \mathcal{S}_\mathcal{I}(G)$ is $\mathcal{I}$-equivalent (Subsection \ref{def:i_equivalent}) to the true graph $G^*$, for $\lambda > 0$ small enough. In case of unknown intervention targets, an additional $-\lambda_R |\mathcal{I}|$ regularization term is added to the score function.

The DCDI algorithm has been tested against IGSP and GIES with known interventions and JCI-PC and UT-IGSP for unknown interventions, showing marginal advantages in terms of structural recovery. As for others continuous optimization methods \cite{zheng2018dags}, the major strength is represented by its scalability: it takes $O(n^3)$, with $n$ the number of variables, to compute the matrix exponential during each training step, making it the only causal discovery algorithm that supports non-linear relationships in the interventional setting in a high-dimensional setting.

%% file: sections/07_evalutation_tuning.tex
\section{Evaluation and Tuning}\label{sec:evaluation}

This section tackles the evaluation and tuning step typical of any practical application. A collection of reference datasets is listed in Table \ref{tab:static_datasets}, both real-world and synthetic generated ones, serving as benchmarking resources for discovery methods. In order to evaluate different solutions resulting from a set of configurations (i.e. hyperparameters) we report  comparison metrics found in the specialized  literature, both in terms of structure and entailed causal statements. Finally, tuning strategies and software packages are explored as support for new developed techniques.

\subsection{Evaluation Datasets}

\paragraph{Cause-Effect Pairs (Tuebingen)}
Ever-growing dataset \cite{mooij2016distinguishing, tagasovska2020distinguishing} designed to benchmark discovery algorithms against bi-variate settings with known ground truth. The latest version reported by the change-log (December 20, 2017) includes 108 pairs. Each pair is composed by a data file with two columns (cause and effect, respectively), a short description of the data sampling procedure, and a 2D scatter plot.

\paragraph{Robotic Manipulation (CausalWorld)}
Simulator \cite{ahmed2020causalworld} for causal structure and transfer learning in a robotic manipulation environment. The environment is a simulation of an open-source robotic platform capable of constructing 3D shapes from a given set of blocks.

\paragraph{Single-Cell Flow Cytometry (Sachs)}
Flow cytometry measurements \cite{sachs2005causal} of 11 proteins and phospholipids. The dataset is split into different experiments, with nine stimulatory or inhibitory interventions. The study compares new learned model against ground truth obtained by reference literature on signaling networks with intervention points.

\paragraph{Single-Cell RNA-Sequencing (Klein)}
Single-cell RNA-sequencing (scRNA-seq) dataset \cite{klein2015droplet} of mouse embryonic stem cells after leukemia inhibitory factor (LIF) withdrawal. The ground truth model is obtained by querying the TRRUST database \cite{han2018trrust} for the related causal relationships.

\paragraph{Single-Cell Gene Expression (Perturb-Seq)}
Measurements of gene expression \cite{dixit2016perturb} composed by 992 observational and 13.435 interventional observations from eight close-to-perfect interventions, each corresponding to a gene deletion using the CRISPR/Cas9 technique applied to bone marrow-derived dendritic cells.

\paragraph{Synthetic Gene Expression (SynTReN)}
Network generator \cite{van2006syntren} that creates synthetic transcriptional regulatory networks. The models are paired with kinetics simulations in order to sample gene expression data that approximate observed experimental data.

\paragraph{Synthetic mRNA Expression (DREAM4)}
The DREAM4 challenge \cite{shannon2021dream4} provides five datasets simulated from five biologically plausible gene regulatory networks with 10 genes \cite{marbach2009generating}. Each dataset is composed by both observational and interventional data sampled by applying small random noise perturbations, single-gene knockdowns and single-gene knockouts, resulting in time-series with unknown interventions.

\begin{table}
    \centering
    \begin{tabular}{|c|c|c|c|} \hline
        Dataset & Source & Type & URL \\ \hline
        Tuebingen \cite{mooij2016distinguishing} & R & O & \href{https://webdav.tuebingen.mpg.de/cause-effect/pairs.zip}{Here} \\ \hline
        CausalWorld \cite{ahmed2020causalworld} & S & I & \href{https://github.com/rr-learning/CausalWorld}{Here} \\ \hline
        Sachs \cite{sachs2005causal} & R & M & \href{https://www.science.org/doi/suppl/10.1126/science.1105809/suppl_file/sachs.som.datasets.zip}{Here} \\ \hline
        Klein \cite{klein2015droplet} & R & M & \href{https://www.sciencedirect.com/sdfe/pdf/download/S0092867415005000/attachments}{Here} \\ \hline
        Perturb-Seq \cite{dixit2016perturb} & R & I & \href{https://www.ncbi.nlm.nih.gov/pmc/articles/PMC5181115/}{Here} \\ \hline
        SynTReN \cite{van2006syntren} & S & O & \href{http://bioinformatics.intec.ugent.be/kmarchal/SynTReN/index.html}{Here} \\ \hline
        DREAM4 \cite{shannon2021dream4} & S & M & \href{https://www.bioconductor.org/packages/release/data/experiment/vignettes/DREAM4/inst/doc/DREAM4.pdf}{Here} \\ \hline
    \end{tabular}
    \caption{Static datasets by source, type and availability. Each dataset originates from (R)eal-world or (S)ynthetic experiments, which may contain (O)bservational, (I)nterventional or (M)ixed data, i.e. both observational and interventional instances.}
    \label{tab:static_datasets}
\end{table}

\subsection{Evaluation Metrics}
In the context of causal discovery, the definitions of true positive (TP), true negative (TN), false positive (FP) and false negative (FN) have the same interpretation of the metrics referred to a binary classifier, which tries to predict the edge orientation.

\paragraph{Adjacency Precision (AP) \& Recall (AR)} A first set of evaluation metrics for graphical models is made of the adjacency precision (AP) and adjacency recall (AR) \cite{scheines2016measurement}. These metrics are computed as the ratio between the number of correctly predicted adjacent vertices over the total predicted ones for AP, and true predicted ones for AR. Formally, once the confusion matrix associated with the presence of edges is computed, the two metrics are defined as follows:

\noindent\begin{minipage}{.5\linewidth}
\begin{equation}
    AP = \frac{TP}{TP + FP}
\end{equation}
\end{minipage}%
\begin{minipage}{.5\linewidth}
\begin{equation}
    AR = \frac{TP}{TP + FN}
\end{equation}
\end{minipage}

\paragraph{Arrowheads Precison (AHP) \& Recall (AHR)} While metrics related to adjacency can deliver insights on the general structure (i.e. skeleton) quality, arrowheads metrics \cite{andrews2019learning, scheines2016measurement} focus on highlighting inferred relationships performance. This class of metrics is particularly useful when there are multiple arrowhead marks that encode different causal statements, such as in PAGs. Here, classical adjacency metrics fail to account for invariant marks that might be interpreted as a head or a tail, overestimating the algorithm performance.

As for adjacency metrics, arrowheads precision (AHP) and recall (AHR) are defined as the ratio between correctly predicted arrowheads over total predicted arrowheads, and correctly predicted arrowheads over true arrowheads:

\noindent\begin{minipage}{.51\linewidth}
\begin{equation}
    AHP = \frac{TP}{TP + FP}
\end{equation}
\end{minipage}%
\begin{minipage}{.51\linewidth}
\begin{equation}
    AHR = \frac{TP}{TP + FN}
\end{equation}
\end{minipage} \\

\noindent where TP, FP and FN refer to the confusion matrix entries computed over the predicted arrowheads, not only the presence/absence of an edge.

\paragraph{Structural Hamming Distance (SHD)} It measures the differences between two graphical models in terms of their edges. Formally, let $G$ and $H$ be two graphs and $\textbf{E}(G, U)$ the symmetric difference between the edge sets $\textbf{E}(G)$ and $\textbf{E}(U)$, the SHD \cite{tsamardinos2006max} counts the number of necessary operations to transform $G$ into $H$:
\begin{equation}
    SHD(G, H) = \sum_{(X, Y), \; X < Y}^{\textbf{V}^2}
    \begin{cases}
        1 & (X, Y) \in \textbf{E}(G, U), \\
        1 & (Y, X) \in \textbf{E}(U, G), \\
        0 & \text{otherwise}.
    \end{cases}
\end{equation}
where the allowed operations consist in addition, deletion and reversal of an edge.

\paragraph{Structural Intervention Distance (SID)} It is a pre-metric defined over the interventional distributions. Formally, the SID \cite{peters2015structural} counts the number of wrongly inferred interventional distributions. This measure relies on the notion of adjustment set \cite{shpitser2008complete} and it is strongly related to the SHD.

\subsection{Parameters Tuning}
Strategies to perform parameters tuning are rarely found in surveys, even tough casual discovery algorithms may have multiple parameters that regulate the search procedure. Here, we report three general and flexible practices described in the specialized literature that can be applied to any technique described so far.

\paragraph{Minimizing Model Complexity (BIC \& AIC)}
A first approach for parameters tuning is related to model complexity. The goal is to find the parameters configuration that minimizes the complexity of the associated causal graph. As a measure of complexity one can rely on the Akaike Information Criterion (AIC) (Equation \ref{eq:aic}) or the Bayesian Information Criterion (BIC)  (Equation \ref{eq:bic}). This tuning strategy is particularly effective when coupled with score-based approaches that are able to exploit the same function, allowing to reuse the intermediate scores for a faster evaluation. The most general form of model complexity minimization is implemented as a grid search over all parameters configurations for the ranges.

\paragraph{Stability Approach to Regularization Selection (StARS)}
The StARS \cite{liu2010stability} approach is based on selecting the parameters configuration that minimizes the graph \emph{instability} when small perturbations are applied to the data. The instability of an edge is the probability of presence of said edge when the causal graph is learned from a subsample of the data (without replacement). Hence, the graph instability of a given parameters' configuration $h$ is the average of the edge instabilities computed w.r.t. $h$. In order to avoid configurations that lead to trivial graphs, e.g. the empty graph or the complete graph, the authors introduce a $\beta$ parameter that acts as a threshold for the acceptable level of instability. In the end, this method measures the sensitivity of a specific parameters' configuration $h$ as a function of the underlying data distribution.

\paragraph{Out-of-sample Causal Tuning (OCT)}
While previous approaches focus on metrics related to the causal structure alone, authors in \cite{Biza2020TuningCD} propose to employ the resulting model for its prediction capabilities, reducing the problem into an evaluation of a predictor. This approach works in a out-of-sample fashion, hence the name \emph{Out-of-sample Causal Tuning} (OCT). The main advantages of such method are i) the lack of parametric assumptions about the distribution of the data, and ii) the generalization to cases where the BIC and AIC scores are not defined, i.e. discrete models with hidden variables.
 
\subsection{Software Packages}
Stable and reliable implementations of discovery methods are fundamental to achieve reproducibility of the experimental results. In the following paragraphs, a list of notable tools is presented.

\paragraph{Causal Discovery Toolbox (CDT)}
The Causal Discovery Toolbox \cite{kalainathan2019causal} is a Python front-end that acts as a bridge between different subpackages, pooling together multiple discovery algorithms. For example, one may find constraint-based algorithms such as PC, Max-Min Parents \& Children (MMPC) \cite{tsamardinos2006max}, score-based algorithms as GES and variants (GIES), and non linear approaches as LiNGAM, Causal Additive Models (CAM) \cite{buhlmann2014cam}, and others.

\paragraph{bnlearn}
The bnlearn \cite{scutari2010bnlearn} package is an R package developed for bayesian inference and structural learning. While both PC and MMPC are implemented, algorithms such as Incremental Association Markov Blanket (IAMB) \cite{tsamardinos2003algorithms} and its variants are present too. Moreover, the underlying implementation is well suited for large scale application due to the optimized support for parallel computing \cite{scutari2017parallel}.

\paragraph{pcalg}
The pcalg \cite{kalish2012pcalg} package is a R utility for causal discovery and causal inference using graphical models. The algorithms provided here are PC and variants (CPC, PC Select), FCI and variants (RFCI \cite{colombo2012learning}, Anytime FCI \cite{spirtes2001anytime}, Adaptive Anytime FCI \cite{colombo2012learning, spirtes2001anytime}, FCI-JCI), GES and variants (AGES \cite{nandy2018highdimensional}, ARGES, GIES) and LiNGAM. Given the wide variety of FCI-based algorithms and the integrated tools for causal inference, this package is particularly well suited for causal insufficient settings.

\paragraph{TETRAD}
While previous packages were intended for command line usage, TETRAD \cite{ramsey2018tetrad} is a causal discovery package developed in Java with a graphical user interface. It follows a pipeline paradigm, where the user can drag \& drop boxes from the side bar and connect them together to form data pipelines. The user can choose from a wide range of options, such as PC (PCStable, CPC \& CPCStable \cite{colombo2013orderindependent}, PCMax), FCI (RFCI, RFCI-BSC \cite{jabbari2017discovery}, GFCI), GES (FGES, FGES-MB, IMaGES), LiNGAM, and others. Given the simplicity of the interface, it is well suited for researchers with limited programming experience.

%% file: sections/08_applications.tex
\section{Practical Applications}\label{sec:applications}

\subsection{Causal Discovery in Economics}

\paragraph{Emissions, Production and Energy Use}
Authors in \cite{addo2021co2} explore the interactions between growth in $CO_2$ emissions, economic production and energy use, both at the global and multi-regional levels over the period 1990–2014. In order to recover the causal relationship between variables, a modified version of the PC algorithm for time-series is used.

Using PC for multi-variate analysis in an economic application lead to a high-level understanding of the influences present in the global economic cycle. Moreover, authors highlight the advantages of applying causal discovery approaches w.r.t. other methodologies, e.g. ``Granger causality'', especially with the selected algorithms. In fact, the PC algorithm is specifically designed to avoid conditioning on irrelevant variables, resulting in larger effect size and lower dimensionality.

The output of the discovery step showed that $CO_2$ emissions, energy and economic activity are linked by a set of non-linear dependencies. At the global level, this graph suggests that a too rapid transition to net-zero emissions in the energy sector may hinder the global economic growth. When the regional level is taken into account, it is shown that regions are fully integrated into the system, which argues for coordinated policies across regions.

\subsection{Causal Discovery in Medicine}

\paragraph{Alzheimer's Pathophysiology}
Researchers in \cite{shen2020challenges} employed data made available by the \emph{Alzheimer's Disease Neuroimaging Initiative} (ADNI) coupled with biological markers and clinical assessment to study the biological mechanism behind the Alzheimer's Disease. Two causal discovery algorithms (FCI and FGES) were compared against the \emph{gold standard} graph retrieved from literature.

Moreover, performing the discovery step alternatively under causal sufficient and insufficient assumptions provides useful hints to isolate the influence of unmeasured external variables, e.g. the presence of a bi-directed edge instead of a directed one. These differences in the recovered structures guide researches during the interpretation of the collected data, and ease the evaluation of competing hypotheses that are consistent with the causal graph.

The methods were executed both with and without trivial background knowledge, e.g. patient's age is not affected by any biomarker. A significant improvement was observed with the addition of the knowledge base. Finally, longitudinal data were included, discovering more edges and removing the incorrect ones. The performance of the constraint-based approach was lower and less stable across the bootstrap samples than the score-based one.

\paragraph{Unmet Treatments in Schizophrenia}
Authors in \cite{miley2021schizophrenia} selected the GFCI algorithm to identify the causes of functional outcomes of patients affected by schizophrenia during the \emph{critical window} for early intervention. The algorithm was applied to the \emph{Recovery After an Initial Schizophrenia Episode Early Treatment Program} (RAISE-ETP) trial at two time-points, i.e. baseline and after 6-months. Social and occupational functioning metrics were derived from the \emph{Quality of Life Scale} (QLS). The  retrieved causal graph was used to build a SCM in order to quantify the magnitude of the effects.

Researchers explicitly selected this approach in order to rule-out direct and indirect effects of latent variables, while taking advantage of the hybrid structure of the algorithm to speed-up the discovery step. In fact, the GFCI algorithm leverages an initial phase that exploits the score-based approach described in FGES, which inherently caches intermediate evaluations of the score function to reduce the execution time.

The estimated effects shed light over the interaction between both social and occupational functioning with the socio-affective capacity, which in turn affects the motivation of the subject. Moreover, an extended analysis of time dependencies revealed several causal cycles over the 6-months time-frame.

\subsection{Causal Discovery in Psychology}

\paragraph{Alcohol Use and Anxiety Disorder}
Psychopathology researchers in \cite{anker2019alcohol} used graphical modeling algorithms to identify causal relationships within and between manifestations of psychiatric disorders. In this context, the GFCI algorithm was employed to identify symptoms that are part of a causal chain of ``mediators''. The main target of the study was to  test whether drinking motivated by the goal of reducing negative affect (i.e. \emph{drinking to cope}, DTC) served as a mediator in comorbid alcohol use and anxiety disorder.

One of the reasons why GFCI is well suited for this scenario is due to the capability of this algorithm to tackle the causally insufficient assumption. This property constitutes a major advantage, enabling researches to explore the effect of potential hidden variables during the assessment of the drinking-anxiety interplay in general.

The resulting graph showed that the most important causal influence of drinking was drinking craving, which was in turn influenced by DTC. However, there was still a degree of ambiguity in the direction of depression's associations with social anxiety and stress, suggesting the possible presence of latent variables.

%% file: sections/09_conclusions_future_work.tex
\section{Conclusions and Discussion}\label{sec:conclusions}

\subsection{A Brief Summary}
Causal inference depends heavily on the construction of a reference model that crystallizes the acquired knowledge. To meet such requirement, causal discovery provides a set of methods that are able to recover a graphical description of the underlying mechanism, exploiting both collected data and prior knowledge. In this work, we presented a list of algorithms, evaluation criteria and software tools, trying to cover a wide range of theoretical and practical scenarios in a coherent and unified manner. Moreover, we compared these resources against challenging problems, such as the presence of unobserved variables, cyclical dependencies, non-linear relationships and unknown interventions, highlighting the strengths and weaknesses of each solution. Finally, we reported a set of parameters tuning strategies and publicly available datasets to explore properly the described techniques and to test new ones.

\subsection{Future Directions}
In terms of opportunities for future extensions, in this contribution we did not explore the implications of applying such method to time-series, which would add complexity. Indeed, the representation of the causal dependencies in time are different from the one expressed in a static scenario and deserves a separated discussion on its own, especially when combined with the other topics introduced during the discussion.

From a more general perspective, there is a set of issues that have yet to be solved in order to shorten the distance between these techniques and real-world applications. For instance, causal discovery in presence of missing data is an under explored field. In fact, imputing missing data when the missingness mechanism is not at random results in sub-optimal solutions due to the introduction of bias component. There have been some recent advances w.r.t. to this topic \mbox{\cite{strobl2018fast, witte2021multiple}}, but these new contributions are restricted to constraint-based approaches only.

Another aspect that is worth mentioning is the application of causal discovery to heterogeneous data \mbox{\cite{huang2020causal}}. The ``heterogeneity'' of the data is referred to the collection process that happens under different observational contexts. Na{\"i}vely pooling different data sources together into a single one results in poor performances. Essentially, this problem stems from the presence of distribution shifts, which arise when context variables differ significantly, e.g. in case of environmental changes. Still, learning a causal representation in the presence of multiple heterogeneous sources is a relevant issue that has yet to be solved.

Finally, there are other scenarios in which causal discovery is limited by the current state-of-the-art, such as non-identical overlapping sets of variables \mbox{\cite{mooij2020joint, triantafillou2015constraint}}, learning from streaming data \mbox{\cite{yu2010causal}} and federated learning \mbox{\cite{gao2021federated}}.

\section*{Acknowledgements} We thank the two anonymous reviewers for their insightful comments. We also thank professor Daniela Besozzi for her thoughtful suggestions.

\section*{Funding} \small{Alessio Zanga was granted a Ph.D. scholarship by F. Hoffmann-La Roche Ltd.}

%% file: main.bbl
\begin{thebibliography}{127}
\expandafter\ifx\csname natexlab\endcsname\relax\def\natexlab#1{#1}\fi
\providecommand{\url}[1]{\texttt{#1}}
\providecommand{\href}[2]{#2}
\providecommand{\path}[1]{#1}
\providecommand{\DOIprefix}{doi:}
\providecommand{\ArXivprefix}{arXiv:}
\providecommand{\URLprefix}{URL: }
\providecommand{\Pubmedprefix}{pmid:}
\providecommand{\doi}[1]{\href{http://dx.doi.org/#1}{\path{#1}}}
\providecommand{\Pubmed}[1]{\href{pmid:#1}{\path{#1}}}
\providecommand{\bibinfo}[2]{#2}
\ifx\xfnm\relax \def\xfnm[#1]{\unskip,\space#1}\fi
\bibitem[{Imbens(2004)}]{imbens2004nonparametric}
\bibinfo{author}{G.~W. Imbens},
\newblock \bibinfo{title}{Nonparametric estimation of average treatment effects
  under exogeneity: A review},
\newblock \bibinfo{journal}{Review of Economics and Statistics}
  \bibinfo{volume}{86} (\bibinfo{year}{2004}) \bibinfo{pages}{4--29}.
\bibitem[{of~the Psychiatric Genomics~Consortium
  et~al.(2013)}]{cross2013identification}
\bibinfo{author}{C.-D.~G. of~the Psychiatric Genomics~Consortium}, et~al.,
\newblock \bibinfo{title}{Identification of risk loci with shared effects on
  five major psychiatric disorders: a genome-wide analysis},
\newblock \bibinfo{journal}{The Lancet} \bibinfo{volume}{381}
  (\bibinfo{year}{2013}) \bibinfo{pages}{1371--1379}.
\bibitem[{Hill(2011)}]{hill2011bayesian}
\bibinfo{author}{J.~L. Hill},
\newblock \bibinfo{title}{Bayesian nonparametric modeling for causal
  inference},
\newblock \bibinfo{journal}{Journal of Computational and Graphical Statistics}
  \bibinfo{volume}{20} (\bibinfo{year}{2011}) \bibinfo{pages}{217--240}.
\bibitem[{Pearl(2018)}]{pearl2018theoretical}
\bibinfo{author}{J.~Pearl},
\newblock \bibinfo{title}{Theoretical impediments to machine learning with
  seven sparks from the causal revolution},
\newblock \bibinfo{journal}{Proceedings of the Eleventh ACM International
  Conference on Web Search and Data Mining}  (\bibinfo{year}{2018}).
\bibitem[{Glymour et~al.(2019)Glymour, Zhang, and Spirtes}]{glymour2019review}
\bibinfo{author}{C.~Glymour}, \bibinfo{author}{K.~Zhang},
  \bibinfo{author}{P.~Spirtes},
\newblock \bibinfo{title}{Review of causal discovery methods based on graphical
  models},
\newblock \bibinfo{journal}{Frontiers in Genetics} \bibinfo{volume}{10}
  (\bibinfo{year}{2019}) \bibinfo{pages}{1--15}.
\bibitem[{Hernán and Robins(2020)}]{hernan2020whatif}
\bibinfo{author}{M.~Hernán}, \bibinfo{author}{J.~Robins},
  \bibinfo{title}{Causal Inference: What If}, \bibinfo{publisher}{Boca Raton:
  Chapman \& Hall/CRC}, \bibinfo{year}{2020}.
\bibitem[{Spirtes et~al.(2000)Spirtes, Glymour, Scheines, and
  Heckerman}]{spirtes2000causation}
\bibinfo{author}{P.~Spirtes}, \bibinfo{author}{C.~N. Glymour},
  \bibinfo{author}{R.~Scheines}, \bibinfo{author}{D.~Heckerman},
  \bibinfo{title}{Causation, prediction, and search}, \bibinfo{publisher}{MIT
  Press}, \bibinfo{year}{2000}.
\bibitem[{Bareinboim et~al.(2022)Bareinboim, Correa, Ibeling, and
  Icard}]{bareinboim20211OP}
\bibinfo{author}{E.~Bareinboim}, \bibinfo{author}{J.~D. Correa},
  \bibinfo{author}{D.~Ibeling}, \bibinfo{author}{T.~F. Icard},
\newblock \bibinfo{title}{On pearl’s hierarchy and the foundations of causal
  inference},
\newblock \bibinfo{journal}{Probabilistic and Causal Inference}
  (\bibinfo{year}{2022}).
\bibitem[{Glymour et~al.(2016)Glymour, Pearl, and Jewell}]{glymour2016causal}
\bibinfo{author}{M.~Glymour}, \bibinfo{author}{J.~Pearl},
  \bibinfo{author}{N.~P. Jewell}, \bibinfo{title}{Causal inference in
  statistics: A primer}, \bibinfo{publisher}{John Wiley \& Sons},
  \bibinfo{year}{2016}.
\bibitem[{Nogueira et~al.(2022)Nogueira, Pugnana, Ruggieri, Pedreschi, and
  Gama}]{nogueira2022methods}
\bibinfo{author}{A.~R. Nogueira}, \bibinfo{author}{A.~Pugnana},
  \bibinfo{author}{S.~Ruggieri}, \bibinfo{author}{D.~Pedreschi},
  \bibinfo{author}{J.~Gama},
\newblock \bibinfo{title}{Methods and tools for causal discovery and causal
  inference},
\newblock \bibinfo{journal}{Wiley Interdisciplinary Reviews: Data Mining and
  Knowledge Discovery}  (\bibinfo{year}{2022}) \bibinfo{pages}{e1449}.
\bibitem[{Guo et~al.(2021)Guo, Cheng, Li, Hahn, and Liu}]{guo2021survey}
\bibinfo{author}{R.~Guo}, \bibinfo{author}{L.~Cheng}, \bibinfo{author}{J.~Li},
  \bibinfo{author}{P.~R. Hahn}, \bibinfo{author}{H.~Liu},
\newblock \bibinfo{title}{A survey of learning causality with data: Problems
  and methods},
\newblock \bibinfo{journal}{ACM Computing Surveys} \bibinfo{volume}{53}
  (\bibinfo{year}{2021}) \bibinfo{pages}{1–37}. \URLprefix
  \url{http://dx.doi.org/10.1145/3397269}. \DOIprefix\doi{10.1145/3397269}.
\bibitem[{Moraffah et~al.(2021)Moraffah, Sheth, Karami, Bhattacharya, Wang,
  Tahir, Raglin, and Liu}]{moraffah2021causal}
\bibinfo{author}{R.~Moraffah}, \bibinfo{author}{P.~Sheth},
  \bibinfo{author}{M.~Karami}, \bibinfo{author}{A.~Bhattacharya},
  \bibinfo{author}{Q.~Wang}, \bibinfo{author}{A.~Tahir},
  \bibinfo{author}{A.~Raglin}, \bibinfo{author}{H.~Liu},
\newblock \bibinfo{title}{Causal inference for time series analysis: Problems,
  methods and evaluation},
\newblock \bibinfo{journal}{Knowledge and Information Systems}
  (\bibinfo{year}{2021}) \bibinfo{pages}{1--45}.
\bibitem[{Malinsky and Danks(2018)}]{malinsky2018causal}
\bibinfo{author}{D.~Malinsky}, \bibinfo{author}{D.~Danks},
\newblock \bibinfo{title}{Causal discovery algorithms: {A} practical guide},
\newblock \bibinfo{journal}{Philosophy Compass} \bibinfo{volume}{13}
  (\bibinfo{year}{2018}) \bibinfo{pages}{e12470}.
\bibitem[{Vowels et~al.(2021)Vowels, Camgoz, and Bowden}]{vowels2021d}
\bibinfo{author}{M.~J. Vowels}, \bibinfo{author}{N.~C. Camgoz},
  \bibinfo{author}{R.~Bowden},
\newblock \bibinfo{title}{D'ya like {DAG}s? {A} survey on structure learning
  and causal discovery},
\newblock \bibinfo{journal}{ACM Computing Surveys (CSUR)}
  (\bibinfo{year}{2021}).
\bibitem[{Nogueira et~al.(2021)Nogueira, Gama, and
  Ferreira}]{nogueira2021causal}
\bibinfo{author}{A.~R. Nogueira}, \bibinfo{author}{J.~Gama},
  \bibinfo{author}{C.~A. Ferreira},
\newblock \bibinfo{title}{Causal discovery in machine learning: Theories and
  applications},
\newblock \bibinfo{journal}{Journal of Dynamics \& Games} \bibinfo{volume}{8}
  (\bibinfo{year}{2021}) \bibinfo{pages}{203--231}.
\bibitem[{Sch{\"o}lkopf et~al.(2021)Sch{\"o}lkopf, Locatello, Bauer, Ke,
  Kalchbrenner, Goyal, and Bengio}]{scholkopf2021toward}
\bibinfo{author}{B.~Sch{\"o}lkopf}, \bibinfo{author}{F.~Locatello},
  \bibinfo{author}{S.~Bauer}, \bibinfo{author}{N.~R. Ke},
  \bibinfo{author}{N.~Kalchbrenner}, \bibinfo{author}{A.~Goyal},
  \bibinfo{author}{Y.~Bengio},
\newblock \bibinfo{title}{Toward causal representation learning},
\newblock \bibinfo{journal}{Proceedings of the IEEE} \bibinfo{volume}{109}
  (\bibinfo{year}{2021}) \bibinfo{pages}{612--634}.
\bibitem[{Pearl(2018)}]{Pearl2018BayesianN}
\bibinfo{author}{J.~Pearl},
\newblock \bibinfo{title}{Bayesian networks},
\newblock in: \bibinfo{booktitle}{Encyclopedia of Social Network Analysis and
  Mining. 2nd Ed.}, \bibinfo{year}{2018}.
\bibitem[{Massmann et~al.(2021)Massmann, Gentine, and
  Runge}]{massmann2021causal}
\bibinfo{author}{A.~Massmann}, \bibinfo{author}{P.~Gentine},
  \bibinfo{author}{J.~Runge},
\newblock \bibinfo{title}{Causal inference for process understanding in earth
  sciences},
\newblock \bibinfo{journal}{arXiv preprint arXiv:2105.00912}
  (\bibinfo{year}{2021}).
\bibitem[{Spirtes and Zhang(2016)}]{spirtes2016causal}
\bibinfo{author}{P.~Spirtes}, \bibinfo{author}{K.~Zhang},
\newblock \bibinfo{title}{Causal discovery and inference: concepts and recent
  methodological advances},
\newblock in: \bibinfo{booktitle}{Applied informatics},
  volume~\bibinfo{volume}{3}, \bibinfo{organization}{SpringerOpen},
  \bibinfo{year}{2016}, pp. \bibinfo{pages}{1--28}.
\bibitem[{Bongers and Mooij(2018)}]{bongers2018random}
\bibinfo{author}{S.~Bongers}, \bibinfo{author}{J.~M. Mooij},
  \bibinfo{title}{From random differential equations to structural causal
  models: the stochastic case}, \bibinfo{year}{2018}.
  \href{http://arxiv.org/abs/1803.08784}{{\tt arXiv:1803.08784}}.
\bibitem[{Rubenstein et~al.(2018)Rubenstein, Bongers, Mooij, and
  Sch{\"o}lkopf}]{rubenstein2018deterministic}
\bibinfo{author}{P.~K. Rubenstein}, \bibinfo{author}{S.~Bongers},
  \bibinfo{author}{J.~M. Mooij}, \bibinfo{author}{B.~Sch{\"o}lkopf},
\newblock \bibinfo{title}{From deterministic odes to dynamic structural causal
  models},
\newblock in: \bibinfo{booktitle}{UAI}, \bibinfo{year}{2018}.
\bibitem[{Shahbazinia et~al.(2021)Shahbazinia, Salehkaleybar, and
  Hashemi}]{shahbazinia2021paralingam}
\bibinfo{author}{A.~Shahbazinia}, \bibinfo{author}{S.~Salehkaleybar},
  \bibinfo{author}{M.~Hashemi},
\newblock \bibinfo{title}{Paralingam: Parallel causal structure learning for
  linear non-gaussian acyclic models},
\newblock \bibinfo{journal}{arXiv preprint arXiv:2109.13993}
  (\bibinfo{year}{2021}).
\bibitem[{Shimizu(2014)}]{shimizu2014lingam}
\bibinfo{author}{S.~Shimizu},
\newblock \bibinfo{title}{Lingam: Non-gaussian methods for estimating causal
  structures},
\newblock \bibinfo{journal}{Behaviormetrika} \bibinfo{volume}{41}
  (\bibinfo{year}{2014}) \bibinfo{pages}{65--98}.
\bibitem[{Bongers et~al.(2021)Bongers, Forré, Peters, and
  Mooij}]{bongers2021foundations}
\bibinfo{author}{S.~Bongers}, \bibinfo{author}{P.~Forré},
  \bibinfo{author}{J.~Peters}, \bibinfo{author}{J.~M. Mooij},
  \bibinfo{title}{Foundations of structural causal models with cycles and
  latent variables}, \bibinfo{year}{2021}.
  \href{http://arxiv.org/abs/1611.06221}{{\tt arXiv:1611.06221}}.
\bibitem[{Mooij and Claassen(2020)}]{mooij2020constraint}
\bibinfo{author}{J.~M. Mooij}, \bibinfo{author}{T.~Claassen},
\newblock \bibinfo{title}{Constraint-based causal discovery using partial
  ancestral graphs in the presence of cycles},
\newblock in: \bibinfo{booktitle}{Conference on Uncertainty in Artificial
  Intelligence}, \bibinfo{organization}{PMLR}, \bibinfo{year}{2020}, pp.
  \bibinfo{pages}{1159--1168}.
\bibitem[{Pearl(1995)}]{pearl1995causal}
\bibinfo{author}{J.~Pearl},
\newblock \bibinfo{title}{Causal diagrams for empirical research},
\newblock \bibinfo{journal}{Biometrika} \bibinfo{volume}{82}
  (\bibinfo{year}{1995}) \bibinfo{pages}{669--688}.
\bibitem[{Verma and Pearl(1990)}]{verma1991equivalence}
\bibinfo{author}{T.~Verma}, \bibinfo{author}{J.~Pearl},
\newblock \bibinfo{title}{Equivalence and synthesis of causal models},
\newblock in: \bibinfo{booktitle}{Proceedings of the Sixth Annual Conference on
  Uncertainty in Artificial Intelligence}, UAI '90,
  \bibinfo{publisher}{Elsevier Science Inc.}, \bibinfo{address}{USA},
  \bibinfo{year}{1990}, p. \bibinfo{pages}{255–270}.
\bibitem[{Mooij et~al.(2020)Mooij, Magliacane, and Claassen}]{mooij2020joint}
\bibinfo{author}{J.~M. Mooij}, \bibinfo{author}{S.~Magliacane},
  \bibinfo{author}{T.~Claassen},
\newblock \bibinfo{title}{Joint causal inference from multiple contexts},
\newblock \bibinfo{journal}{J. Mach. Learn. Res.} \bibinfo{volume}{21}
  (\bibinfo{year}{2020}) \bibinfo{pages}{99:1--99:108}.
\bibitem[{Yang et~al.(2018)Yang, Katcoff, and Uhler}]{yang2018characterizing}
\bibinfo{author}{K.~Yang}, \bibinfo{author}{A.~Katcoff},
  \bibinfo{author}{C.~Uhler},
\newblock \bibinfo{title}{Characterizing and learning equivalence classes of
  causal {DAG}s under interventions},
\newblock in: \bibinfo{booktitle}{International Conference on Machine
  Learning}, \bibinfo{organization}{PMLR}, \bibinfo{year}{2018}, pp.
  \bibinfo{pages}{5541--5550}.
\bibitem[{Andersson et~al.(1997)Andersson, Madigan, and
  Perlman}]{andersson1997characterization}
\bibinfo{author}{S.~A. Andersson}, \bibinfo{author}{D.~Madigan},
  \bibinfo{author}{M.~D. Perlman},
\newblock \bibinfo{title}{A characterization of {M}arkov equivalence classes
  for acyclic digraphs},
\newblock \bibinfo{journal}{The Annals of Statistics} \bibinfo{volume}{25}
  (\bibinfo{year}{1997}) \bibinfo{pages}{505--541}.
\bibitem[{Meek(2013)}]{meek2013causal}
\bibinfo{author}{C.~Meek},
\newblock \bibinfo{title}{Causal inference and causal explanation with
  background knowledge},
\newblock \bibinfo{journal}{arXiv preprint arXiv:1302.4972}
  (\bibinfo{year}{2013}).
\bibitem[{Kocaoglu et~al.(2017)Kocaoglu, Shanmugam, and
  Bareinboim}]{kocaoglu2017experimental}
\bibinfo{author}{M.~Kocaoglu}, \bibinfo{author}{K.~Shanmugam},
  \bibinfo{author}{E.~Bareinboim},
\newblock \bibinfo{title}{Experimental design for learning causal graphs with
  latent variables},
\newblock in: \bibinfo{booktitle}{Proceedings of the 31st International
  Conference on Neural Information Processing Systems}, NIPS'17,
  \bibinfo{publisher}{Curran Associates Inc.}, \bibinfo{address}{Red Hook, NY,
  USA}, \bibinfo{year}{2017}, p. \bibinfo{pages}{7021–7031}.
\bibitem[{Forré and Mooij(2017)}]{forre2017markov}
\bibinfo{author}{P.~Forré}, \bibinfo{author}{J.~M. Mooij},
  \bibinfo{title}{Markov properties for graphical models with cycles and latent
  variables}, \bibinfo{year}{2017}. \href{http://arxiv.org/abs/1710.08775}{{\tt
  arXiv:1710.08775}}.
\bibitem[{Zhang(2008)}]{zhang2008completeness}
\bibinfo{author}{J.~Zhang},
\newblock \bibinfo{title}{On the completeness of orientation rules for causal
  discovery in the presence of latent confounders and selection bias},
\newblock \bibinfo{journal}{Artificial Intelligence} \bibinfo{volume}{172}
  (\bibinfo{year}{2008}) \bibinfo{pages}{1873--1896}.
\bibitem[{Richardson and Spirtes(2002)}]{richardson2002ancestral}
\bibinfo{author}{T.~Richardson}, \bibinfo{author}{P.~Spirtes},
\newblock \bibinfo{title}{Ancestral graph {M}arkov models},
\newblock \bibinfo{journal}{The Annals of Statistics} \bibinfo{volume}{30}
  (\bibinfo{year}{2002}) \bibinfo{pages}{962--1030}.
\bibitem[{Drton and Richardson(2004)}]{drton2012iterative}
\bibinfo{author}{M.~Drton}, \bibinfo{author}{T.~S. Richardson},
\newblock \bibinfo{title}{Iterative conditional fitting for gaussian ancestral
  graph models},
\newblock in: \bibinfo{booktitle}{UAI}, \bibinfo{year}{2004}.
\bibitem[{Peters et~al.(2017)Peters, Janzing, and
  Sch{\"o}lkopf}]{peters2017elements}
\bibinfo{author}{J.~Peters}, \bibinfo{author}{D.~Janzing},
  \bibinfo{author}{B.~Sch{\"o}lkopf}, \bibinfo{title}{Elements of causal
  inference: foundations and learning algorithms}, \bibinfo{publisher}{The MIT
  Press}, \bibinfo{year}{2017}.
\bibitem[{Shimizu and Blöbaum(2020)}]{shimizu2020semiparametric}
\bibinfo{author}{S.~Shimizu}, \bibinfo{author}{P.~Blöbaum},
  \bibinfo{title}{Recent Advances in Semi-Parametric Methods for Causal
  Discovery}, \bibinfo{year}{2020}, pp. \bibinfo{pages}{111--130}. \URLprefix
  \url{https://onlinelibrary.wiley.com/doi/abs/10.1002/9781119523024.ch5}.
  \DOIprefix\doi{https://doi.org/10.1002/9781119523024.ch5}.
  \href{http://arxiv.org/abs/https://onlinelibrary.wiley.com/doi/pdf/10.1002/9781119523024.ch5}{{\tt
  arXiv:https://onlinelibrary.wiley.com/doi/pdf/10.1002/9781119523024.ch5}}.
\bibitem[{Colombo and Maathuis(2013)}]{colombo2013orderindependent}
\bibinfo{author}{D.~Colombo}, \bibinfo{author}{M.~H. Maathuis},
  \bibinfo{title}{Order-independent constraint-based causal structure
  learning}, \bibinfo{year}{2013}. \href{http://arxiv.org/abs/1211.3295}{{\tt
  arXiv:1211.3295}}.
\bibitem[{Alonso-Barba et~al.(2013)Alonso-Barba, G{\'a}mez, Puerta
  et~al.}]{alonso2013scaling}
\bibinfo{author}{J.~I. Alonso-Barba}, \bibinfo{author}{J.~A. G{\'a}mez},
  \bibinfo{author}{J.~M. Puerta}, et~al.,
\newblock \bibinfo{title}{Scaling up the greedy equivalence search algorithm by
  constraining the search space of equivalence classes},
\newblock \bibinfo{journal}{International {J}ournal of {A}pproximate
  {R}easoning} \bibinfo{volume}{54} (\bibinfo{year}{2013})
  \bibinfo{pages}{429--451}.
\bibitem[{Ramsey et~al.(2017)Ramsey, Glymour, Sanchez-Romero, and
  Glymour}]{ramsey2017million}
\bibinfo{author}{J.~Ramsey}, \bibinfo{author}{M.~Glymour},
  \bibinfo{author}{R.~Sanchez-Romero}, \bibinfo{author}{C.~Glymour},
\newblock \bibinfo{title}{A million variables and more: the fast greedy
  equivalence search algorithm for learning high-dimensional graphical causal
  models, with an application to functional magnetic resonance images},
\newblock \bibinfo{journal}{International {J}ournal of {D}ata {S}cience and
  {A}nalytics} \bibinfo{volume}{3} (\bibinfo{year}{2017})
  \bibinfo{pages}{121--129}.
\bibitem[{Nandy et~al.(2018)Nandy, Hauser, and
  Maathuis}]{nandy2018highdimensional}
\bibinfo{author}{P.~Nandy}, \bibinfo{author}{A.~Hauser}, \bibinfo{author}{M.~H.
  Maathuis},
\newblock \bibinfo{title}{High-dimensional consistency in score-based and
  hybrid structure learning},
\newblock \bibinfo{journal}{The Annals of Statistics}  (\bibinfo{year}{2018}).
\bibitem[{Ogarrio et~al.(2016)Ogarrio, Spirtes, and Ramsey}]{ogarrio2016hybrid}
\bibinfo{author}{J.~M. Ogarrio}, \bibinfo{author}{P.~Spirtes},
  \bibinfo{author}{J.~Ramsey},
\newblock \bibinfo{title}{A hybrid causal search algorithm for latent variable
  models},
\newblock in: \bibinfo{booktitle}{Conference on Probabilistic Graphical
  Models}, \bibinfo{organization}{PMLR}, \bibinfo{year}{2016}, pp.
  \bibinfo{pages}{368--379}.
\bibitem[{Cai et~al.(2018)Cai, Qiao, Zhang, Zhang, and Hao}]{cai2018causal}
\bibinfo{author}{R.~Cai}, \bibinfo{author}{J.~Qiao},
  \bibinfo{author}{K.~Zhang}, \bibinfo{author}{Z.~Zhang},
  \bibinfo{author}{Z.~Hao},
\newblock \bibinfo{title}{Causal discovery from discrete data using hidden
  compact representation},
\newblock \bibinfo{journal}{Advances in {N}eural {I}nformation {P}rocessing
  {S}ystems} \bibinfo{volume}{32} (\bibinfo{year}{2018})
  \bibinfo{pages}{2671--2679}.
\bibitem[{Tagasovska et~al.(2020)Tagasovska, Chavez-Demoulin, and
  Vatter}]{tagasovska2020distinguishing}
\bibinfo{author}{N.~Tagasovska}, \bibinfo{author}{V.~Chavez-Demoulin},
  \bibinfo{author}{T.~Vatter},
\newblock \bibinfo{title}{Distinguishing cause from effect using quantiles:
  Bivariate quantile causal discovery},
\newblock in: \bibinfo{booktitle}{International Conference on Machine
  Learning}, \bibinfo{organization}{PMLR}, \bibinfo{year}{2020}, pp.
  \bibinfo{pages}{9311--9323}.
\bibitem[{Hoyer et~al.(2006)Hoyer, Shimizu, and Kerminen}]{hoyer2006estimation}
\bibinfo{author}{P.~O. Hoyer}, \bibinfo{author}{S.~Shimizu},
  \bibinfo{author}{A.~J. Kerminen},
\newblock \bibinfo{title}{Estimation of linear, non-gaussian causal models in
  the presence of confounding latent variables},
\newblock in: \bibinfo{booktitle}{Probabilistic Graphical Models},
  \bibinfo{year}{2006}.
\bibitem[{Zheng et~al.(2018)Zheng, Aragam, Ravikumar, and Xing}]{zheng2018dags}
\bibinfo{author}{X.~Zheng}, \bibinfo{author}{B.~Aragam},
  \bibinfo{author}{P.~Ravikumar}, \bibinfo{author}{E.~P. Xing},
\newblock \bibinfo{title}{D{AG}s with no tears: Continuous optimization for
  structure learning},
\newblock \bibinfo{journal}{arXiv preprint arXiv:1803.01422}
  (\bibinfo{year}{2018}).
\bibitem[{Richardson(2013)}]{richardson2013discovery}
\bibinfo{author}{T.~S. Richardson},
\newblock \bibinfo{title}{A discovery algorithm for directed cyclic graphs},
\newblock \bibinfo{journal}{arXiv preprint arXiv:1302.3599}
  (\bibinfo{year}{2013}).
\bibitem[{Lacerda et~al.(2008)Lacerda, Spirtes, Ramsey, and
  Hoyer}]{lacerda2012discovering}
\bibinfo{author}{G.~Lacerda}, \bibinfo{author}{P.~L. Spirtes},
  \bibinfo{author}{J.~Ramsey}, \bibinfo{author}{P.~O. Hoyer},
\newblock \bibinfo{title}{Discovering cyclic causal models by independent
  components analysis},
\newblock in: \bibinfo{booktitle}{UAI}, \bibinfo{year}{2008}.
\bibitem[{Hyttinen et~al.(2017)Hyttinen, Saikko, J{\"a}rvisalo
  et~al.}]{hyttinen2017core}
\bibinfo{author}{A.~Hyttinen}, \bibinfo{author}{P.~Saikko},
  \bibinfo{author}{M.~J{\"a}rvisalo}, et~al.,
\newblock \bibinfo{title}{A core-guided approach to learning optimal causal
  graphs},
\newblock in: \bibinfo{booktitle}{Proceedings of the 26th International Joint
  Conference on Artificial Intelligence (IJCAI 2017)},
  \bibinfo{organization}{International Joint Conferences on Artificial
  Intelligence}, \bibinfo{year}{2017}.
\bibitem[{Rantanen et~al.(2020)Rantanen, Hyttinen, and
  J{\"a}rvisalo}]{rantanen2020discovering}
\bibinfo{author}{K.~Rantanen}, \bibinfo{author}{A.~Hyttinen},
  \bibinfo{author}{M.~J{\"a}rvisalo},
\newblock \bibinfo{title}{Discovering causal graphs with cycles and latent
  confounders: An exact branch-and-bound approach},
\newblock \bibinfo{journal}{International Journal of Approximate Reasoning}
  \bibinfo{volume}{117} (\bibinfo{year}{2020}) \bibinfo{pages}{29--49}.
\bibitem[{Forr{\'e} and Mooij(2018)}]{forre2018constraint}
\bibinfo{author}{P.~Forr{\'e}}, \bibinfo{author}{J.~M. Mooij},
\newblock \bibinfo{title}{Constraint-based causal discovery for non-linear
  structural causal models with cycles and latent confounders},
\newblock \bibinfo{journal}{arXiv preprint arXiv:1807.03024}
  (\bibinfo{year}{2018}).
\bibitem[{Hauser and B{\"u}hlmann(2012)}]{hauser2012characterization}
\bibinfo{author}{A.~Hauser}, \bibinfo{author}{P.~B{\"u}hlmann},
\newblock \bibinfo{title}{Characterization and greedy learning of
  interventional {M}arkov equivalence classes of directed acyclic graphs},
\newblock \bibinfo{journal}{The Journal of Machine Learning Research}
  \bibinfo{volume}{13} (\bibinfo{year}{2012}) \bibinfo{pages}{2409--2464}.
\bibitem[{Squires et~al.(2020)Squires, Wang, and
  Uhler}]{squires2020permutationbased}
\bibinfo{author}{C.~Squires}, \bibinfo{author}{Y.~Wang},
  \bibinfo{author}{C.~Uhler}, \bibinfo{title}{Permutation-based causal
  structure learning with unknown intervention targets}, \bibinfo{year}{2020}.
  \href{http://arxiv.org/abs/1910.09007}{{\tt arXiv:1910.09007}}.
\bibitem[{Jaber et~al.(2020)Jaber, Kocaoglu, Shanmugam, and
  Bareinboim}]{jaber2020advances}
\bibinfo{author}{A.~Jaber}, \bibinfo{author}{M.~Kocaoglu},
  \bibinfo{author}{K.~Shanmugam}, \bibinfo{author}{E.~Bareinboim},
\newblock \bibinfo{title}{Causal discovery from soft interventions with unknown
  targets: Characterization and learning},
\newblock in: \bibinfo{editor}{H.~Larochelle}, \bibinfo{editor}{M.~Ranzato},
  \bibinfo{editor}{R.~Hadsell}, \bibinfo{editor}{M.~F. Balcan},
  \bibinfo{editor}{H.~Lin} (Eds.), \bibinfo{booktitle}{Advances in Neural
  Information Processing Systems}, volume~\bibinfo{volume}{33},
  \bibinfo{publisher}{Curran Associates, Inc.}, \bibinfo{year}{2020}, pp.
  \bibinfo{pages}{9551--9561}. \URLprefix
  \url{https://proceedings.neurips.cc/paper/2020/file/6cd9313ed34ef58bad3fdd504355e72c-Paper.pdf}.
\bibitem[{Rothenh\"{a}usler et~al.(2015)Rothenh\"{a}usler, Heinze, Peters, and
  Meinshausen}]{rothenhausler2015backshift}
\bibinfo{author}{D.~Rothenh\"{a}usler}, \bibinfo{author}{C.~Heinze},
  \bibinfo{author}{J.~Peters}, \bibinfo{author}{N.~Meinshausen},
\newblock \bibinfo{title}{Backshift: Learning causal cyclic graphs from unknown
  shift interventions},
\newblock in: \bibinfo{editor}{C.~Cortes}, \bibinfo{editor}{N.~Lawrence},
  \bibinfo{editor}{D.~Lee}, \bibinfo{editor}{M.~Sugiyama},
  \bibinfo{editor}{R.~Garnett} (Eds.), \bibinfo{booktitle}{Advances in Neural
  Information Processing Systems}, volume~\bibinfo{volume}{28},
  \bibinfo{publisher}{Curran Associates, Inc.}, \bibinfo{year}{2015}.
  \URLprefix
  \url{https://proceedings.neurips.cc/paper/2015/file/92262bf907af914b95a0fc33c3f33bf6-Paper.pdf}.
\bibitem[{Rantanen et~al.(2020)Rantanen, Hyttinen, and
  J{\"a}rvisalo}]{rantanen2020learning}
\bibinfo{author}{K.~Rantanen}, \bibinfo{author}{A.~Hyttinen},
  \bibinfo{author}{M.~J{\"a}rvisalo},
\newblock \bibinfo{title}{Learning optimal cyclic causal graphs from
  interventional data},
\newblock in: \bibinfo{booktitle}{International Conference on Probabilistic
  Graphical Models}, \bibinfo{organization}{PMLR}, \bibinfo{year}{2020}, pp.
  \bibinfo{pages}{365--376}.
\bibitem[{Brouillard et~al.(2020)Brouillard, Lachapelle, Lacoste,
  Lacoste-Julien, and Drouin}]{brouillard2020differentiable}
\bibinfo{author}{P.~Brouillard}, \bibinfo{author}{S.~Lachapelle},
  \bibinfo{author}{A.~Lacoste}, \bibinfo{author}{S.~Lacoste-Julien},
  \bibinfo{author}{A.~Drouin}, \bibinfo{title}{Differentiable causal discovery
  from interventional data}, \bibinfo{year}{2020}.
  \href{http://arxiv.org/abs/2007.01754}{{\tt arXiv:2007.01754}}.
\bibitem[{Castillo et~al.(2012)Castillo, Gutierrez, and
  Hadi}]{castillo2012expert}
\bibinfo{author}{E.~Castillo}, \bibinfo{author}{J.~M. Gutierrez},
  \bibinfo{author}{A.~S. Hadi}, \bibinfo{title}{Expert systems and
  probabilistic network models}, \bibinfo{publisher}{Springer Science \&
  Business Media}, \bibinfo{year}{2012}.
\bibitem[{Koller and Friedman(2009)}]{koller2009probabilistic}
\bibinfo{author}{D.~Koller}, \bibinfo{author}{N.~Friedman},
  \bibinfo{title}{Probabilistic graphical models: principles and techniques},
  \bibinfo{publisher}{MIT {P}ress}, \bibinfo{year}{2009}.
\bibitem[{Andrews et~al.(2019)Andrews, Ramsey, and
  Cooper}]{andrews2019learning}
\bibinfo{author}{B.~Andrews}, \bibinfo{author}{J.~Ramsey},
  \bibinfo{author}{G.~F. Cooper},
\newblock \bibinfo{title}{Learning high-dimensional directed acyclic graphs
  with mixed data-types},
\newblock in: \bibinfo{booktitle}{The 2019 ACM SIGKDD Workshop on Causal
  Discovery}, \bibinfo{organization}{PMLR}, \bibinfo{year}{2019}, pp.
  \bibinfo{pages}{4--21}.
\bibitem[{Tsagris et~al.(2018)Tsagris, Borboudakis, Lagani, and
  Tsamardinos}]{tsagris2018constraint}
\bibinfo{author}{M.~Tsagris}, \bibinfo{author}{G.~Borboudakis},
  \bibinfo{author}{V.~Lagani}, \bibinfo{author}{I.~Tsamardinos},
\newblock \bibinfo{title}{Constraint-based causal discovery with mixed data},
\newblock \bibinfo{journal}{International {J}ournal of {D}ata {S}cience and
  {A}nalytics} \bibinfo{volume}{6} (\bibinfo{year}{2018})
  \bibinfo{pages}{19--30}.
\bibitem[{Le et~al.(2019)Le, Hoang, Li, Liu, Liu, and Hu}]{thucduy2019fastpc}
\bibinfo{author}{T.~D. Le}, \bibinfo{author}{T.~Hoang},
  \bibinfo{author}{J.~Li}, \bibinfo{author}{L.~Liu}, \bibinfo{author}{H.~Liu},
  \bibinfo{author}{S.~Hu},
\newblock \bibinfo{title}{A fast {PC} algorithm for high dimensional causal
  discovery with multi-core pcs},
\newblock \bibinfo{journal}{IEEE/ACM Transactions on Computational Biology and
  Bioinformatics} \bibinfo{volume}{16} (\bibinfo{year}{2019})
  \bibinfo{pages}{1483–1495}. \URLprefix
  \url{http://dx.doi.org/10.1109/TCBB.2016.2591526}.
  \DOIprefix\doi{10.1109/tcbb.2016.2591526}.
\bibitem[{Li and Fan(2020)}]{li2020nonparametric}
\bibinfo{author}{C.~Li}, \bibinfo{author}{X.~Fan},
\newblock \bibinfo{title}{On nonparametric conditional independence tests for
  continuous variables},
\newblock \bibinfo{journal}{Wiley Interdisciplinary Reviews: Computational
  Statistics} \bibinfo{volume}{12} (\bibinfo{year}{2020})
  \bibinfo{pages}{e1489}.
\bibitem[{Spirtes et~al.(2013)Spirtes, Meek, and
  Richardson}]{spirtes2013causal}
\bibinfo{author}{P.~L. Spirtes}, \bibinfo{author}{C.~Meek},
  \bibinfo{author}{T.~S. Richardson},
\newblock \bibinfo{title}{Causal inference in the presence of latent variables
  and selection bias}  (\bibinfo{year}{2013}).
  \href{http://arxiv.org/abs/1302.4983}{{\tt arXiv:1302.4983}}.
\bibitem[{Lee et~al.(2020)Lee, Correa, and Bareinboim}]{lee2020generalized}
\bibinfo{author}{S.~Lee}, \bibinfo{author}{J.~Correa},
  \bibinfo{author}{E.~Bareinboim},
\newblock \bibinfo{title}{Generalized transportability: Synthesis of
  experiments from heterogeneous domains},
\newblock in: \bibinfo{booktitle}{Proceedings of the 34th AAAI Conference on
  Artificial Intelligence}, \bibinfo{publisher}{AAAI Press},
  \bibinfo{address}{New York, NY}, \bibinfo{year}{2020}.
\bibitem[{Chickering(2002)}]{chickering2002optimal}
\bibinfo{author}{D.~M. Chickering},
\newblock \bibinfo{title}{Optimal structure identification with greedy search},
\newblock \bibinfo{journal}{Journal of {M}achine {L}earning {R}esearch}
  \bibinfo{volume}{3} (\bibinfo{year}{2002}) \bibinfo{pages}{507--554}.
\bibitem[{Akaike(1974)}]{akaike1974criterion}
\bibinfo{author}{H.~Akaike},
\newblock \bibinfo{title}{A new look at the statistical model identification},
\newblock \bibinfo{journal}{IEEE Transactions on Automatic Control}
  \bibinfo{volume}{19} (\bibinfo{year}{1974}) \bibinfo{pages}{716--723}.
  \DOIprefix\doi{10.1109/TAC.1974.1100705}.
\bibitem[{Schwarz(1978)}]{schwarz1978estimating}
\bibinfo{author}{G.~Schwarz},
\newblock \bibinfo{title}{Estimating the dimension of a model},
\newblock \bibinfo{journal}{The {A}nnals of {S}tatistics}
  (\bibinfo{year}{1978}) \bibinfo{pages}{461--464}.
\bibitem[{Geiger and Heckerman(1994)}]{geiger1994learning}
\bibinfo{author}{D.~Geiger}, \bibinfo{author}{D.~Heckerman},
\newblock \bibinfo{title}{Learning {G}aussian networks},
\newblock in: \bibinfo{booktitle}{Uncertainty Proceedings 1994},
  \bibinfo{publisher}{Elsevier}, \bibinfo{year}{1994}, pp.
  \bibinfo{pages}{235--243}.
\bibitem[{Scutari(2016)}]{scutari2016empirical}
\bibinfo{author}{M.~Scutari},
\newblock \bibinfo{title}{An empirical-{B}ayes score for discrete {B}ayesian
  networks},
\newblock in: \bibinfo{booktitle}{Conference on {P}robabilistic {G}raphical
  {M}odels}, \bibinfo{organization}{PMLR}, \bibinfo{year}{2016}, pp.
  \bibinfo{pages}{438--448}.
\bibitem[{Meek(1997)}]{meek1997graphical}
\bibinfo{author}{C.~Meek}, \bibinfo{title}{Graphical Models: Selecting causal
  and statistical models}, Ph.D. thesis, PhD thesis, Carnegie Mellon
  University, \bibinfo{year}{1997}.
\bibitem[{Rissanen(1978)}]{rissanen1978modeling}
\bibinfo{author}{J.~Rissanen},
\newblock \bibinfo{title}{Modeling by shortest data description},
\newblock \bibinfo{journal}{Automatica} \bibinfo{volume}{14}
  (\bibinfo{year}{1978}) \bibinfo{pages}{465--471}.
\bibitem[{Janzing and Sch{\"o}lkopf(2010)}]{janzing2010causal}
\bibinfo{author}{D.~Janzing}, \bibinfo{author}{B.~Sch{\"o}lkopf},
\newblock \bibinfo{title}{Causal inference using the algorithmic {M}arkov
  condition},
\newblock \bibinfo{journal}{IEEE Transactions on Information Theory}
  \bibinfo{volume}{56} (\bibinfo{year}{2010}) \bibinfo{pages}{5168--5194}.
\bibitem[{Stegle et~al.(2010)Stegle, Janzing, Zhang, Mooij, and
  Sch{\"o}lkopf}]{stegle2010probabilistic}
\bibinfo{author}{O.~Stegle}, \bibinfo{author}{D.~Janzing},
  \bibinfo{author}{K.~Zhang}, \bibinfo{author}{J.~M. Mooij},
  \bibinfo{author}{B.~Sch{\"o}lkopf},
\newblock \bibinfo{title}{Probabilistic latent variable models for
  distinguishing between cause and effect},
\newblock \bibinfo{journal}{Advances in {N}eural {I}nformation {P}rocessing
  {S}ystems} \bibinfo{volume}{23} (\bibinfo{year}{2010})
  \bibinfo{pages}{1687--1695}.
\bibitem[{Comon(1994)}]{comon1994independent}
\bibinfo{author}{P.~Comon},
\newblock \bibinfo{title}{Independent component analysis, a new concept?},
\newblock \bibinfo{journal}{Signal {P}rocessing} \bibinfo{volume}{36}
  (\bibinfo{year}{1994}) \bibinfo{pages}{287--314}.
\bibitem[{Tsamardinos et~al.(2006)Tsamardinos, Brown, and
  Aliferis}]{tsamardinos2006max}
\bibinfo{author}{I.~Tsamardinos}, \bibinfo{author}{L.~E. Brown},
  \bibinfo{author}{C.~F. Aliferis},
\newblock \bibinfo{title}{The max-min hill-climbing {B}ayesian network
  structure learning algorithm},
\newblock \bibinfo{journal}{Machine {L}earning} \bibinfo{volume}{65}
  (\bibinfo{year}{2006}) \bibinfo{pages}{31--78}.
\bibitem[{Niinimaki and Parviainen(2012)}]{niinimaki2012local}
\bibinfo{author}{T.~Niinimaki}, \bibinfo{author}{P.~Parviainen},
\newblock \bibinfo{title}{Local structure discovery in bayesian networks},
\newblock in: \bibinfo{booktitle}{UAI}, \bibinfo{year}{2012}.
\bibitem[{Natori et~al.(2015)Natori, Uto, Nishiyama, Kawano, and
  Ueno}]{natori2015constraint}
\bibinfo{author}{K.~Natori}, \bibinfo{author}{M.~Uto},
  \bibinfo{author}{Y.~Nishiyama}, \bibinfo{author}{S.~Kawano},
  \bibinfo{author}{M.~Ueno},
\newblock \bibinfo{title}{Constraint-based learning {B}ayesian networks using
  {B}ayes factor},
\newblock in: \bibinfo{booktitle}{Workshop on Advanced Methodologies for
  Bayesian Networks}, \bibinfo{organization}{Springer}, \bibinfo{year}{2015},
  pp. \bibinfo{pages}{15--31}.
\bibitem[{Scutari et~al.(2019)Scutari, Graafland, and
  Guti{\'e}rrez}]{scutari2019learns}
\bibinfo{author}{M.~Scutari}, \bibinfo{author}{C.~E. Graafland},
  \bibinfo{author}{J.~M. Guti{\'e}rrez},
\newblock \bibinfo{title}{Who learns better {B}ayesian network structures:
  Accuracy and speed of structure learning algorithms},
\newblock \bibinfo{journal}{International Journal of Approximate Reasoning}
  \bibinfo{volume}{115} (\bibinfo{year}{2019}) \bibinfo{pages}{235--253}.
\bibitem[{Spirtes(2010)}]{spirtes2010introduction}
\bibinfo{author}{P.~Spirtes},
\newblock \bibinfo{title}{Introduction to causal inference},
\newblock \bibinfo{journal}{Journal of Machine Learning Research}
  \bibinfo{volume}{11} (\bibinfo{year}{2010}) \bibinfo{pages}{1643--1662}.
  \URLprefix \url{http://jmlr.org/papers/v11/spirtes10a.html}.
\bibitem[{Berry(1984)}]{berry1984nonrecursive}
\bibinfo{author}{W.~D. Berry}, \bibinfo{title}{Nonrecursive causal models},
  \bibinfo{number}{37}, \bibinfo{publisher}{Sage}, \bibinfo{year}{1984}.
\bibitem[{Nagase and Kano(2017)}]{nagase2017identifiability}
\bibinfo{author}{M.~Nagase}, \bibinfo{author}{Y.~Kano},
\newblock \bibinfo{title}{Identifiability of nonrecursive structural equation
  models},
\newblock \bibinfo{journal}{Statistics \& Probability Letters}
  \bibinfo{volume}{122} (\bibinfo{year}{2017}) \bibinfo{pages}{109--117}.
\bibitem[{Spirtes(2013)}]{spirtes2013directed}
\bibinfo{author}{P.~L. Spirtes}, \bibinfo{title}{Directed cyclic graphical
  representations of feedback models}, \bibinfo{year}{2013}.
  \href{http://arxiv.org/abs/1302.4982}{{\tt arXiv:1302.4982}}.
\bibitem[{Hyttinen et~al.(2014)Hyttinen, Eberhardt, and
  J\"{a}rvisalo}]{hyttinen2014constraint}
\bibinfo{author}{A.~Hyttinen}, \bibinfo{author}{F.~Eberhardt},
  \bibinfo{author}{M.~J\"{a}rvisalo},
\newblock \bibinfo{title}{Constraint-based causal discovery: Conflict
  resolution with answer set programming},
\newblock in: \bibinfo{booktitle}{Proceedings of the Thirtieth Conference on
  Uncertainty in Artificial Intelligence}, UAI'14, \bibinfo{publisher}{AUAI
  Press}, \bibinfo{address}{Arlington, Virginia, USA}, \bibinfo{year}{2014}, p.
  \bibinfo{pages}{340–349}.
\bibitem[{Magliacane et~al.(2017)Magliacane, Claassen, and
  Mooij}]{magliacane2017ancestral}
\bibinfo{author}{S.~Magliacane}, \bibinfo{author}{T.~Claassen},
  \bibinfo{author}{J.~M. Mooij}, \bibinfo{title}{Ancestral causal inference},
  \bibinfo{year}{2017}. \href{http://arxiv.org/abs/1606.07035}{{\tt
  arXiv:1606.07035}}.
\bibitem[{Pearl and Mackenzie(2018)}]{pearl2018why}
\bibinfo{author}{J.~Pearl}, \bibinfo{author}{D.~Mackenzie}, \bibinfo{title}{The
  Book of Why: The New Science of Cause and Effect}, \bibinfo{edition}{1st}
  ed., \bibinfo{publisher}{Basic Books, Inc.}, \bibinfo{address}{USA},
  \bibinfo{year}{2018}.
\bibitem[{Shpitser and Pearl(2008)}]{shpitser2008complete}
\bibinfo{author}{I.~Shpitser}, \bibinfo{author}{J.~Pearl},
\newblock \bibinfo{title}{Complete identification methods for the causal
  hierarchy.},
\newblock \bibinfo{journal}{Journal of Machine Learning Research}
  \bibinfo{volume}{9} (\bibinfo{year}{2008}).
\bibitem[{Markowetz et~al.(2005)Markowetz, Grossmann, and
  Spang}]{markowetz2005probabilistic}
\bibinfo{author}{F.~Markowetz}, \bibinfo{author}{S.~Grossmann},
  \bibinfo{author}{R.~Spang},
\newblock \bibinfo{title}{Probabilistic soft interventions in conditional
  {G}aussian networks},
\newblock in: \bibinfo{editor}{R.~G. Cowell}, \bibinfo{editor}{Z.~Ghahramani}
  (Eds.), \bibinfo{booktitle}{Proceedings of the Tenth International Workshop
  on Artificial Intelligence and Statistics}, volume~\bibinfo{volume}{R5} of
  \textit{\bibinfo{series}{Proceedings of Machine Learning Research}},
  \bibinfo{publisher}{PMLR}, \bibinfo{year}{2005}, pp.
  \bibinfo{pages}{214--221}. \URLprefix
  \url{https://proceedings.mlr.press/r5/markowetz05a.html},
  \bibinfo{note}{reissued by PMLR on 30 March 2021.}
\bibitem[{Tian and Pearl(2013)}]{tian2013causal}
\bibinfo{author}{J.~Tian}, \bibinfo{author}{J.~Pearl}, \bibinfo{title}{Causal
  discovery from changes}, \bibinfo{year}{2013}.
  \href{http://arxiv.org/abs/1301.2312}{{\tt arXiv:1301.2312}}.
\bibitem[{Kocaoglu et~al.(2019)Kocaoglu, Jaber, Shanmugam, and
  Bareinboim}]{kocaoglu2019advances}
\bibinfo{author}{M.~Kocaoglu}, \bibinfo{author}{A.~Jaber},
  \bibinfo{author}{K.~Shanmugam}, \bibinfo{author}{E.~Bareinboim},
\newblock \bibinfo{title}{Characterization and learning of causal graphs with
  latent variables from soft interventions},
\newblock in: \bibinfo{editor}{H.~Wallach}, \bibinfo{editor}{H.~Larochelle},
  \bibinfo{editor}{A.~Beygelzimer}, \bibinfo{editor}{F.~d\textquotesingle
  Alch\'{e}-Buc}, \bibinfo{editor}{E.~Fox}, \bibinfo{editor}{R.~Garnett}
  (Eds.), \bibinfo{booktitle}{Advances in Neural Information Processing
  Systems}, volume~\bibinfo{volume}{32}, \bibinfo{publisher}{Curran Associates,
  Inc.}, \bibinfo{year}{2019}. \URLprefix
  \url{https://proceedings.neurips.cc/paper/2019/file/c3d96fbd5b1b45096ff04c04038fff5d-Paper.pdf}.
\bibitem[{Eberhardt et~al.(2012)Eberhardt, Glymour, and
  Scheines}]{eberhardt2012number}
\bibinfo{author}{F.~Eberhardt}, \bibinfo{author}{C.~Glymour},
  \bibinfo{author}{R.~Scheines}, \bibinfo{title}{On the number of experiments
  sufficient and in the worst case necessary to identify all causal relations
  among n variables}, \bibinfo{year}{2012}.
  \href{http://arxiv.org/abs/1207.1389}{{\tt arXiv:1207.1389}}.
\bibitem[{Hu et~al.(2014)Hu, Li, and Vetta}]{hu2014randomized}
\bibinfo{author}{H.~Hu}, \bibinfo{author}{Z.~Li}, \bibinfo{author}{A.~R.
  Vetta},
\newblock \bibinfo{title}{Randomized experimental design for causal graph
  discovery},
\newblock in: \bibinfo{editor}{Z.~Ghahramani}, \bibinfo{editor}{M.~Welling},
  \bibinfo{editor}{C.~Cortes}, \bibinfo{editor}{N.~Lawrence},
  \bibinfo{editor}{K.~Weinberger} (Eds.), \bibinfo{booktitle}{Advances in
  Neural Information Processing Systems}, volume~\bibinfo{volume}{27},
  \bibinfo{publisher}{Curran Associates, Inc.}, \bibinfo{year}{2014}.
  \URLprefix
  \url{https://proceedings.neurips.cc/paper/2014/file/e53a0a2978c28872a4505bdb51db06dc-Paper.pdf}.
\bibitem[{Solus et~al.(2021)Solus, Wang, and Uhler}]{solus2021consistency}
\bibinfo{author}{L.~Solus}, \bibinfo{author}{Y.~Wang},
  \bibinfo{author}{C.~Uhler}, \bibinfo{title}{Consistency guarantees for greedy
  permutation-based causal inference algorithms}, \bibinfo{year}{2021}.
  \href{http://arxiv.org/abs/1702.03530}{{\tt arXiv:1702.03530}}.
\bibitem[{Mooij et~al.(2016)Mooij, Peters, Janzing, Zscheischler, and
  Sch{\"o}lkopf}]{mooij2016distinguishing}
\bibinfo{author}{J.~M. Mooij}, \bibinfo{author}{J.~Peters},
  \bibinfo{author}{D.~Janzing}, \bibinfo{author}{J.~Zscheischler},
  \bibinfo{author}{B.~Sch{\"o}lkopf},
\newblock \bibinfo{title}{Distinguishing cause from effect using observational
  data: methods and benchmarks},
\newblock \bibinfo{journal}{The Journal of Machine Learning Research}
  \bibinfo{volume}{17} (\bibinfo{year}{2016}) \bibinfo{pages}{1103--1204}.
\bibitem[{Ahmed et~al.(2020)Ahmed, Träuble, Goyal, Neitz, Bengio, Schölkopf,
  Wüthrich, and Bauer}]{ahmed2020causalworld}
\bibinfo{author}{O.~Ahmed}, \bibinfo{author}{F.~Träuble},
  \bibinfo{author}{A.~Goyal}, \bibinfo{author}{A.~Neitz},
  \bibinfo{author}{Y.~Bengio}, \bibinfo{author}{B.~Schölkopf},
  \bibinfo{author}{M.~Wüthrich}, \bibinfo{author}{S.~Bauer},
  \bibinfo{title}{Causal{W}orld: A robotic manipulation benchmark for causal
  structure and transfer learning}, \bibinfo{year}{2020}.
  \href{http://arxiv.org/abs/2010.04296}{{\tt arXiv:2010.04296}}.
\bibitem[{Sachs et~al.(2005)Sachs, Perez, Pe'er, Lauffenburger, and
  Nolan}]{sachs2005causal}
\bibinfo{author}{K.~Sachs}, \bibinfo{author}{O.~Perez},
  \bibinfo{author}{D.~Pe'er}, \bibinfo{author}{D.~A. Lauffenburger},
  \bibinfo{author}{G.~P. Nolan},
\newblock \bibinfo{title}{Causal protein-signaling networks derived from
  multiparameter single-cell data},
\newblock \bibinfo{journal}{Science} \bibinfo{volume}{308}
  (\bibinfo{year}{2005}) \bibinfo{pages}{523--529}.
\bibitem[{Klein et~al.(2015)Klein, Mazutis, Akartuna, Tallapragada, Veres, Li,
  Peshkin, Weitz, and Kirschner}]{klein2015droplet}
\bibinfo{author}{A.~M. Klein}, \bibinfo{author}{L.~Mazutis},
  \bibinfo{author}{I.~Akartuna}, \bibinfo{author}{N.~Tallapragada},
  \bibinfo{author}{A.~Veres}, \bibinfo{author}{V.~Li},
  \bibinfo{author}{L.~Peshkin}, \bibinfo{author}{D.~A. Weitz},
  \bibinfo{author}{M.~W. Kirschner},
\newblock \bibinfo{title}{Droplet barcoding for single-cell transcriptomics
  applied to embryonic stem cells},
\newblock \bibinfo{journal}{Cell} \bibinfo{volume}{161} (\bibinfo{year}{2015})
  \bibinfo{pages}{1187--1201}.
\bibitem[{Han et~al.(2018)Han, Cho, Lee, Yun, Kim, Bae, Yang, Kim, Lee, Kim
  et~al.}]{han2018trrust}
\bibinfo{author}{H.~Han}, \bibinfo{author}{J.-W. Cho},
  \bibinfo{author}{S.~Lee}, \bibinfo{author}{A.~Yun}, \bibinfo{author}{H.~Kim},
  \bibinfo{author}{D.~Bae}, \bibinfo{author}{S.~Yang}, \bibinfo{author}{C.~Y.
  Kim}, \bibinfo{author}{M.~Lee}, \bibinfo{author}{E.~Kim}, et~al.,
\newblock \bibinfo{title}{T{RRUST} v2: an expanded reference database of human
  and mouse transcriptional regulatory interactions},
\newblock \bibinfo{journal}{Nucleic {A}cids {R}esearch} \bibinfo{volume}{46}
  (\bibinfo{year}{2018}) \bibinfo{pages}{D380--D386}.
\bibitem[{Dixit et~al.(2016)Dixit, Parnas, Li, Chen, Fulco, Jerby-Arnon,
  Marjanovic, Dionne, Burks, Raychowdhury et~al.}]{dixit2016perturb}
\bibinfo{author}{A.~Dixit}, \bibinfo{author}{O.~Parnas},
  \bibinfo{author}{B.~Li}, \bibinfo{author}{J.~Chen}, \bibinfo{author}{C.~P.
  Fulco}, \bibinfo{author}{L.~Jerby-Arnon}, \bibinfo{author}{N.~D. Marjanovic},
  \bibinfo{author}{D.~Dionne}, \bibinfo{author}{T.~Burks},
  \bibinfo{author}{R.~Raychowdhury}, et~al.,
\newblock \bibinfo{title}{Perturb-seq: dissecting molecular circuits with
  scalable single-cell {RNA} profiling of pooled genetic screens},
\newblock \bibinfo{journal}{Cell} \bibinfo{volume}{167} (\bibinfo{year}{2016})
  \bibinfo{pages}{1853--1866}.
\bibitem[{Van~den Bulcke et~al.(2006)Van~den Bulcke, Van~Leemput, Naudts, van
  Remortel, Ma, Verschoren, De~Moor, and Marchal}]{van2006syntren}
\bibinfo{author}{T.~Van~den Bulcke}, \bibinfo{author}{K.~Van~Leemput},
  \bibinfo{author}{B.~Naudts}, \bibinfo{author}{P.~van Remortel},
  \bibinfo{author}{H.~Ma}, \bibinfo{author}{A.~Verschoren},
  \bibinfo{author}{B.~De~Moor}, \bibinfo{author}{K.~Marchal},
\newblock \bibinfo{title}{Syn{TR}e{N}: a generator of synthetic gene expression
  data for design and analysis of structure learning algorithms},
\newblock \bibinfo{journal}{BMC {B}ioinformatics} \bibinfo{volume}{7}
  (\bibinfo{year}{2006}) \bibinfo{pages}{1--12}.
\bibitem[{Shannon(2021)}]{shannon2021dream4}
\bibinfo{author}{P.~Shannon}, \bibinfo{title}{Dream4: Synthetic expression data
  for gene regulatory network inference from the 2009 {DREAM4} challenge},
  \bibinfo{year}{2021}. \bibinfo{note}{R package version 1.30.0}.
\bibitem[{Marbach et~al.(2009)Marbach, Schaffter, Mattiussi, and
  Floreano}]{marbach2009generating}
\bibinfo{author}{D.~Marbach}, \bibinfo{author}{T.~Schaffter},
  \bibinfo{author}{C.~Mattiussi}, \bibinfo{author}{D.~Floreano},
\newblock \bibinfo{title}{Generating realistic in silico gene networks for
  performance assessment of reverse engineering methods},
\newblock \bibinfo{journal}{Journal of {C}omputational {B}iology}
  \bibinfo{volume}{16} (\bibinfo{year}{2009}) \bibinfo{pages}{229--239}.
\bibitem[{Scheines and Ramsey(2016)}]{scheines2016measurement}
\bibinfo{author}{R.~Scheines}, \bibinfo{author}{J.~Ramsey},
\newblock \bibinfo{title}{Measurement error and causal discovery},
\newblock in: \bibinfo{booktitle}{CEUR workshop proceedings}, volume
  \bibinfo{volume}{1792}, \bibinfo{organization}{NIH Public Access},
  \bibinfo{year}{2016}, p.~\bibinfo{pages}{1}.
\bibitem[{Peters and B{\"u}hlmann(2015)}]{peters2015structural}
\bibinfo{author}{J.~Peters}, \bibinfo{author}{P.~B{\"u}hlmann},
\newblock \bibinfo{title}{Structural intervention distance for evaluating
  causal graphs},
\newblock \bibinfo{journal}{Neural {C}omputation} \bibinfo{volume}{27}
  (\bibinfo{year}{2015}) \bibinfo{pages}{771--799}.
\bibitem[{Liu et~al.(2010)Liu, Roeder, and Wasserman}]{liu2010stability}
\bibinfo{author}{H.~Liu}, \bibinfo{author}{K.~Roeder}, \bibinfo{author}{L.~A.
  Wasserman},
\newblock \bibinfo{title}{Stability approach to regularization selection
  (stars) for high dimensional graphical models},
\newblock \bibinfo{journal}{Advances in neural information processing systems}
  \bibinfo{volume}{24 2} (\bibinfo{year}{2010}) \bibinfo{pages}{1432--1440}.
\bibitem[{Biza et~al.(2020)Biza, Tsamardinos, and
  Triantafillou}]{Biza2020TuningCD}
\bibinfo{author}{K.~Biza}, \bibinfo{author}{I.~Tsamardinos},
  \bibinfo{author}{S.~Triantafillou},
\newblock \bibinfo{title}{Tuning causal discovery algorithms},
\newblock in: \bibinfo{editor}{M.~Jaeger}, \bibinfo{editor}{T.~D. Nielsen}
  (Eds.), \bibinfo{booktitle}{Proceedings of the 10th International Conference
  on Probabilistic Graphical Models}, volume \bibinfo{volume}{138} of
  \textit{\bibinfo{series}{Proceedings of Machine Learning Research}},
  \bibinfo{publisher}{PMLR}, \bibinfo{year}{2020}, pp. \bibinfo{pages}{17--28}.
  \URLprefix \url{https://proceedings.mlr.press/v138/biza20a.html}.
\bibitem[{Kalainathan and Goudet(2019)}]{kalainathan2019causal}
\bibinfo{author}{D.~Kalainathan}, \bibinfo{author}{O.~Goudet},
  \bibinfo{title}{Causal discovery toolbox: Uncover causal relationships in
  {P}ython}, \bibinfo{year}{2019}. \href{http://arxiv.org/abs/1903.02278}{{\tt
  arXiv:1903.02278}}.
\bibitem[{B{\"u}hlmann et~al.(2014)B{\"u}hlmann, Peters, and
  Ernest}]{buhlmann2014cam}
\bibinfo{author}{P.~B{\"u}hlmann}, \bibinfo{author}{J.~Peters},
  \bibinfo{author}{J.~Ernest},
\newblock \bibinfo{title}{C{AM}: Causal additive models, high-dimensional order
  search and penalized regression},
\newblock \bibinfo{journal}{The Annals of Statistics} \bibinfo{volume}{42}
  (\bibinfo{year}{2014}) \bibinfo{pages}{2526--2556}.
\bibitem[{Scutari(2010)}]{scutari2010bnlearn}
\bibinfo{author}{M.~Scutari},
\newblock \bibinfo{title}{Learning {B}ayesian networks with the {bnlearn} {R}
  package},
\newblock \bibinfo{journal}{Journal of {S}tatistical {S}oftware}
  \bibinfo{volume}{35} (\bibinfo{year}{2010}) \bibinfo{pages}{1--22}.
  \DOIprefix\doi{10.18637/jss.v035.i03}.
\bibitem[{Tsamardinos et~al.(2003)Tsamardinos, Aliferis, Statnikov, and
  Statnikov}]{tsamardinos2003algorithms}
\bibinfo{author}{I.~Tsamardinos}, \bibinfo{author}{C.~F. Aliferis},
  \bibinfo{author}{A.~R. Statnikov}, \bibinfo{author}{E.~Statnikov},
\newblock \bibinfo{title}{Algorithms for large scale {M}arkov blanket
  discovery.},
\newblock in: \bibinfo{booktitle}{FLAIRS conference},
  volume~\bibinfo{volume}{2}, \bibinfo{organization}{St. Augustine, FL},
  \bibinfo{year}{2003}, pp. \bibinfo{pages}{376--380}.
\bibitem[{Scutari(2017)}]{scutari2017parallel}
\bibinfo{author}{M.~Scutari},
\newblock \bibinfo{title}{Bayesian network constraint-based structure learning
  algorithms: Parallel and optimized implementations in the {bnlearn} {R}
  package},
\newblock \bibinfo{journal}{Journal of Statistical Software}
  \bibinfo{volume}{77} (\bibinfo{year}{2017}) \bibinfo{pages}{1--20}.
  \DOIprefix\doi{10.18637/jss.v077.i02}.
\bibitem[{Kalisch et~al.(2012)Kalisch, M\"achler, Colombo, Maathuis, and
  B\"uhlmann}]{kalish2012pcalg}
\bibinfo{author}{M.~Kalisch}, \bibinfo{author}{M.~M\"achler},
  \bibinfo{author}{D.~Colombo}, \bibinfo{author}{M.~H. Maathuis},
  \bibinfo{author}{P.~B\"uhlmann},
\newblock \bibinfo{title}{Causal inference using graphical models with the {R}
  package {pcalg}},
\newblock \bibinfo{journal}{Journal of Statistical Software}
  \bibinfo{volume}{47} (\bibinfo{year}{2012}) \bibinfo{pages}{1--26}.
  \DOIprefix\doi{10.18637/jss.v047.i11}.
\bibitem[{Colombo et~al.(2012)Colombo, Maathuis, Kalisch, and
  Richardson}]{colombo2012learning}
\bibinfo{author}{D.~Colombo}, \bibinfo{author}{M.~H. Maathuis},
  \bibinfo{author}{M.~Kalisch}, \bibinfo{author}{T.~S. Richardson},
\newblock \bibinfo{title}{Learning high-dimensional directed acyclic graphs
  with latent and selection variables},
\newblock \bibinfo{journal}{The Annals of Statistics} \bibinfo{volume}{40}
  (\bibinfo{year}{2012}). \URLprefix \url{http://dx.doi.org/10.1214/11-AOS940}.
  \DOIprefix\doi{10.1214/11-aos940}.
\bibitem[{Spirtes(2001)}]{spirtes2001anytime}
\bibinfo{author}{P.~Spirtes},
\newblock \bibinfo{title}{An anytime algorithm for causal inference},
\newblock in: \bibinfo{booktitle}{International Workshop on Artificial
  Intelligence and Statistics}, \bibinfo{organization}{PMLR},
  \bibinfo{year}{2001}, pp. \bibinfo{pages}{278--285}.
\bibitem[{Ramsey et~al.(2018)Ramsey, Zhang, Glymour, Romero, Huang,
  Ebert-Uphoff, Samarasinghe, Barnes, and Glymour}]{ramsey2018tetrad}
\bibinfo{author}{J.~D. Ramsey}, \bibinfo{author}{K.~Zhang},
  \bibinfo{author}{M.~Glymour}, \bibinfo{author}{R.~S. Romero},
  \bibinfo{author}{B.~Huang}, \bibinfo{author}{I.~Ebert-Uphoff},
  \bibinfo{author}{S.~Samarasinghe}, \bibinfo{author}{E.~A. Barnes},
  \bibinfo{author}{C.~Glymour},
\newblock \bibinfo{title}{T{ETRAD}—a toolbox for causal discovery},
\newblock in: \bibinfo{booktitle}{8th International Workshop on Climate
  Informatics}, \bibinfo{year}{2018}.
\bibitem[{Jabbari et~al.(2017)Jabbari, Ramsey, Spirtes, and
  Cooper}]{jabbari2017discovery}
\bibinfo{author}{F.~Jabbari}, \bibinfo{author}{J.~Ramsey},
  \bibinfo{author}{P.~Spirtes}, \bibinfo{author}{G.~Cooper},
\newblock \bibinfo{title}{Discovery of causal models that contain latent
  variables through {B}ayesian scoring of independence constraints},
\newblock in: \bibinfo{booktitle}{Joint European Conference on Machine Learning
  and Knowledge Discovery in Databases}, \bibinfo{organization}{Springer},
  \bibinfo{year}{2017}, pp. \bibinfo{pages}{142--157}.
\bibitem[{Addo et~al.(2021)Addo, Manibialoa, and McIsaac}]{addo2021co2}
\bibinfo{author}{P.~M. Addo}, \bibinfo{author}{C.~Manibialoa},
  \bibinfo{author}{F.~McIsaac},
\newblock \bibinfo{title}{Exploring nonlinearity on the {CO2} emissions,
  economic production and energy use nexus: A causal discovery approach},
\newblock \bibinfo{journal}{Energy Reports} \bibinfo{volume}{7}
  (\bibinfo{year}{2021}) \bibinfo{pages}{6196--6204}. \URLprefix
  \url{https://www.sciencedirect.com/science/article/pii/S2352484721008313}.
  \DOIprefix\doi{https://doi.org/10.1016/j.egyr.2021.09.026}.
\bibitem[{Shen et~al.(2020)Shen, Ma, Vemuri, and Simon}]{shen2020challenges}
\bibinfo{author}{X.~Shen}, \bibinfo{author}{S.~Ma},
  \bibinfo{author}{P.~Vemuri}, \bibinfo{author}{G.~Simon},
\newblock \bibinfo{title}{Challenges and opportunities with causal discovery
  algorithms: application to {A}lzheimer’s pathophysiology},
\newblock \bibinfo{journal}{Scientific {R}eports} \bibinfo{volume}{10}
  (\bibinfo{year}{2020}) \bibinfo{pages}{1--12}.
\bibitem[{Miley et~al.(2021)Miley, Meyer-Kalos, Ma, Bond, Kummerfeld, and
  Vinogradov}]{miley2021schizophrenia}
\bibinfo{author}{K.~Miley}, \bibinfo{author}{P.~Meyer-Kalos},
  \bibinfo{author}{S.~Ma}, \bibinfo{author}{D.~J. Bond},
  \bibinfo{author}{E.~Kummerfeld}, \bibinfo{author}{S.~Vinogradov},
\newblock \bibinfo{title}{Causal pathways to social and occupational
  functioning in the first episode of schizophrenia: uncovering unmet treatment
  needs},
\newblock \bibinfo{journal}{Psychological Medicine}  (\bibinfo{year}{2021})
  \bibinfo{pages}{1–9}. \DOIprefix\doi{10.1017/S0033291721003780}.
\bibitem[{Anker et~al.(2019)Anker, Kummerfeld, Rix, Burwell, and
  Kushner}]{anker2019alcohol}
\bibinfo{author}{J.~J. Anker}, \bibinfo{author}{E.~Kummerfeld},
  \bibinfo{author}{A.~Rix}, \bibinfo{author}{S.~J. Burwell},
  \bibinfo{author}{M.~G. Kushner},
\newblock \bibinfo{title}{Causal network modeling of the determinants of
  drinking behavior in comorbid alcohol use and anxiety disorder},
\newblock \bibinfo{journal}{Alcoholism: Clinical and Experimental Research}
  \bibinfo{volume}{43} (\bibinfo{year}{2019}) \bibinfo{pages}{91--97}.
  \URLprefix \url{https://onlinelibrary.wiley.com/doi/abs/10.1111/acer.13914}.
  \DOIprefix\doi{https://doi.org/10.1111/acer.13914}.
  \href{http://arxiv.org/abs/https://onlinelibrary.wiley.com/doi/pdf/10.1111/acer.13914}{{\tt
  arXiv:https://onlinelibrary.wiley.com/doi/pdf/10.1111/acer.13914}}.
\bibitem[{Strobl et~al.(2018)Strobl, Visweswaran, and Spirtes}]{strobl2018fast}
\bibinfo{author}{E.~V. Strobl}, \bibinfo{author}{S.~Visweswaran},
  \bibinfo{author}{P.~L. Spirtes},
\newblock \bibinfo{title}{Fast causal inference with non-random missingness by
  test-wise deletion},
\newblock \bibinfo{journal}{International {J}ournal of {D}ata {S}cience and
  {A}nalytics} \bibinfo{volume}{6} (\bibinfo{year}{2018})
  \bibinfo{pages}{47--62}.
\bibitem[{Witte et~al.(2021)Witte, Foraita, and Didelez}]{witte2021multiple}
\bibinfo{author}{J.~Witte}, \bibinfo{author}{R.~Foraita},
  \bibinfo{author}{V.~Didelez},
\newblock \bibinfo{title}{Multiple imputation and test-wise deletion for causal
  discovery with incomplete cohort data},
\newblock \bibinfo{journal}{arXiv preprint arXiv:2108.13331}
  (\bibinfo{year}{2021}).
\bibitem[{Huang et~al.(2020)Huang, Zhang, Zhang, Ramsey, Sanchez-Romero,
  Glymour, and Sch{\"o}lkopf}]{huang2020causal}
\bibinfo{author}{B.~Huang}, \bibinfo{author}{K.~Zhang},
  \bibinfo{author}{J.~Zhang}, \bibinfo{author}{J.~D. Ramsey},
  \bibinfo{author}{R.~Sanchez-Romero}, \bibinfo{author}{C.~Glymour},
  \bibinfo{author}{B.~Sch{\"o}lkopf},
\newblock \bibinfo{title}{Causal discovery from heterogeneous/nonstationary
  data.},
\newblock \bibinfo{journal}{Journal of Machine Learning Research}
  \bibinfo{volume}{21} (\bibinfo{year}{2020}) \bibinfo{pages}{1--53}.
\bibitem[{Triantafillou and Tsamardinos(2015)}]{triantafillou2015constraint}
\bibinfo{author}{S.~Triantafillou}, \bibinfo{author}{I.~Tsamardinos},
\newblock \bibinfo{title}{Constraint-based causal discovery from multiple
  interventions over overlapping variable sets},
\newblock \bibinfo{journal}{The Journal of Machine Learning Research}
  \bibinfo{volume}{16} (\bibinfo{year}{2015}) \bibinfo{pages}{2147--2205}.
\bibitem[{Yu et~al.(2010)Yu, Wu, Wang, and Ding}]{yu2010causal}
\bibinfo{author}{K.~Yu}, \bibinfo{author}{X.~Wu}, \bibinfo{author}{H.~Wang},
  \bibinfo{author}{W.~Ding},
\newblock \bibinfo{title}{Causal discovery from streaming features},
\newblock in: \bibinfo{booktitle}{2010 IEEE International Conference on Data
  Mining}, \bibinfo{organization}{IEEE}, \bibinfo{year}{2010}, pp.
  \bibinfo{pages}{1163--1168}.
\bibitem[{Gao et~al.(2021)Gao, Chen, Shen, Liu, Gong, and
  Bondell}]{gao2021federated}
\bibinfo{author}{E.~Gao}, \bibinfo{author}{J.~Chen}, \bibinfo{author}{L.~Shen},
  \bibinfo{author}{T.~Liu}, \bibinfo{author}{M.~Gong},
  \bibinfo{author}{H.~Bondell},
\newblock \bibinfo{title}{Federated causal discovery},
\newblock \bibinfo{journal}{arXiv preprint arXiv:2112.03555}
  (\bibinfo{year}{2021}).

\end{thebibliography}
